\documentclass[10pt,journal,compsoc]{IEEEtran}
\ifCLASSOPTIONcompsoc
  \usepackage[nocompress]{cite}
\else
  \usepackage{cite}
\fi
\ifCLASSINFOpdf
\else
\fi

\usepackage{bm}
\usepackage{amssymb}
\usepackage{latexsym}
\usepackage{dsfont}
\usepackage{multirow}
\usepackage{amsfonts}
\usepackage{amsmath}
\usepackage{algorithm}
\usepackage{algorithmic}
\usepackage{graphicx}
\usepackage{hyperref}
\usepackage{color}
\usepackage{soul} 

\def\vp{{\bm{p}}}

\def\vu{{\bm{u}}}

\def\vx{{\bm{x}}}

\def\mA{{\bm{A}}}

\def\mD{{\bm{D}}}

\def\mL{{\bm{L}}}

\def\mP{{\bm{P}}}

\def\mU{{\bm{U}}}

\def\mLambda{{\bm{\Lambda}}}

\DeclareMathAlphabet{\mathsfit}{\encodingdefault}{\sfdefault}{m}{sl}
\SetMathAlphabet{\mathsfit}{bold}{\encodingdefault}{\sfdefault}{bx}{n}

\def\gE{{\mathcal{E}}}

\def\gG{{\mathcal{G}}}

\def\gV{{\mathcal{V}}}

\newcommand{\red}[1] {\textcolor{black}{{#1}}}

\newcommand{\new}[1] {\textcolor{black}{{#1}}}

\def\ie{{\it i.e.}}
\def\eg{{\it e.g.}}
\def\et{{\it et al.}}

\begin{document}
%
\title{Point Cloud Attacks in Graph Spectral Domain: When 3D Geometry Meets Graph Signal Processing}

\author{Daizong~Liu,~\IEEEmembership{Student~Member,~IEEE},
        ~Wei~Hu,~\IEEEmembership{Senior~Member,~IEEE},
        and~Xin~Li,~\IEEEmembership{Fellow,~IEEE}
\IEEEcompsocitemizethanks{\IEEEcompsocthanksitem D. Liu and W. Hu are with Wangxuan Institute of Computer Technology, Peking University, No. 128, Zhongguancun North Street, Beijing, China. E-mail: dzliu@stu.pku.edu.cn, forhuwei@pku.edu.cn. 
\IEEEcompsocthanksitem Xin Li is with the Department of Computer Science, University at Albany, Albany, NY 12222 USA. E-mail: xli48@albany.edu.
\IEEEcompsocthanksitem Corresponding author: Wei Hu. This work was supported by National Natural Science Foundation of China (61972009).}}

%
%

\markboth{Journal of \LaTeX\ Class Files,~Vol.~14, No.~8, August~2015}%
{Shell \MakeLowercase{\textit{et al.}}: Bare Demo of IEEEtran.cls for Computer Society Journals}
%



\IEEEtitleabstractindextext{%
\begin{abstract}
With the increasing attention in various 3D safety-critical applications, point cloud learning models have been shown to be vulnerable to adversarial attacks. Although existing 3D attack methods achieve high success rates, they delve into the data space with point-wise perturbation, which may neglect the geometric characteristics. Instead, we propose point cloud attacks from a new perspective---the graph spectral domain attack, aiming to perturb graph transform coefficients in the spectral domain that correspond to varying certain geometric structures. Specifically, leveraging on graph signal processing, we first adaptively transform the coordinates of points onto the spectral domain via graph Fourier transform (GFT) for compact representation. Then, we analyze the influence of different spectral bands on the geometric structure, based on which we propose to perturb the GFT coefficients via a learnable graph spectral filter. Considering the low-frequency components mainly contribute to the rough shape of the 3D object, we further introduce a low-frequency constraint to limit perturbations within imperceptible high-frequency components. Finally, the adversarial point cloud is generated by transforming the perturbed spectral representation back to the data domain via the inverse GFT. Experimental results demonstrate the effectiveness of the proposed attack in terms of both the imperceptibility and attack success rates.
\end{abstract}

}

\maketitle

\IEEEdisplaynontitleabstractindextext

%
\IEEEpeerreviewmaketitle

\IEEEraisesectionheading{\section{Introduction}
\label{sec:introduction}}

\IEEEPARstart{D}{eep} Neural Networks (DNNs) have shown to be vulnerable to adversarial examples \cite{goodfellow2014explaining,szegedy2013intriguing}, which add visually indistinguishable perturbations to network inputs but lead to incorrect prediction results. 
Significant progress has been made in adversarial attacks on 2D images, where many methods \cite{dong2018boosting,madry2017towards,kurakin2016adversarial,tu2019autozoom} learn to add imperceptible pixel-wise noise in the spatial or feature domain.
Nevertheless, adversarial attacks on 3D depth or point-cloud data are still relatively underexplored. With the maturity of depth sensors, 3D point clouds have received increasing attention in various applications such as autonomous driving \cite{chen2017multi} and medical data analysis \cite{singh20203d}. Similarly to their 2D counterparts, deep learning models trained for point clouds are often vulnerable to adversarial perturbations, increasing the risk in such safety-critical applications.

Existing 3D point cloud attacks \cite{xiang2019generating,wicker2019robustness,zhang2019adversarial,zheng2019pointcloud,tsai2020robust,zhao2020isometry,zhou2020lg,hamdi2020advpc} have been developed in the {\it data space}. 
Given a point cloud, some of them \cite{xiang2019generating,zhang2019adversarial,wicker2019robustness,zheng2019pointcloud} generally employ the gradient search method to identify the most critical points in local regions and modify (add or remove) them to distort the most representative features for misclassification. 
Recently, more attempts \cite{wen2020geometry,tsai2020robust,carlini2017towards,hamdi2020advpc,liu2019extending,ma2020efficient,zhang2019defense} follow the C\&W framework \cite{goodfellow2014explaining} to learn to perturb the xyz coordinates of all points by gradient optimization in a learnable and end-to-end manner. 
Although the above two types of attack methods in the {\it data space} achieve high success rates, 
\red{
it is difficult for existing attacks to properly preserve the geometric characteristics of point clouds, with outliers and uneven distributions likely to be introduced.
Geometric-aware loss functions are needed for such approaches to keep the original geometric structure, \textit{e.g.}, curvature loss \cite{wen2020geometry}, which are, however, not always effective enough to measure and preserve general geometric contexts during the gradient backpropagation as a point cloud is often of complicated structure. }
\red{
Moreover, benign point clouds generally have geometric characteristics such as piecewise smoothness \cite{hu2021overview,hu2014multiresolution,chao2015edge}, which exhibit slowly varying underlying surfaces separated by sharp edges. However, directly perturbing points along the xyz directions or modifying local points in the data domain would often deform the geometric structure and may also result in outliers and uneven distribution.}

To address the above limitations, an ideal point cloud attack method \red{would characterize and preserve 3D geometric structures during the perturbation process. 
In general, signals can be compactly represented in the spectral domain.
For example, images are often transformed into the Discrete Cosine Transform (DCT) domain for compression and processing \cite{choi2020task,li2018learning}.
However, different from 2D images that generally contain fine-grained texture, 3D geometric shapes mainly consist of object contours.
Therefore, directly applying 2D spectral methods into the 3D domain is not suitable. 
Besides,}
unlike images supported on regular grids, point clouds reside in {\it irregular} domains with no ordering of points, which hinders the deployment of traditional transforms such as the DCT. 
Fortunately, graphs serve as a natural representation of irregular point clouds, which is accurate and structure-adaptive \cite{hu2021overview}. 
With an appropriately constructed graph that captures the underlying structure well, the graph Fourier transforms (GFT) \cite{hammond2011wavelets} 
\red{have been proven to be approximately optimal in compact representation of geometric data \cite{hu2014multiresolution,zhang2012analyzing}.  
As we build a K-Nearest-Neighbor graph that captures the correlation among points, the corresponding GFT 
will lead to a compact representation of geometric data including point clouds in the spectral domain} \cite{shen2010edge,hu2012depth,hu2014multiresolution,hu2015intra,zhang2014point,xu2020predictive}, which \red{is effective in comprehending the shape information of the whole point cloud. 
To be specific, the graph spectral domain
explicitly reveals the geometric structures from basic shapes to fine details via different frequency bands \cite{hu2021overview}.
This}
inspires new insights and understanding for processing the geometry of point clouds. 

\begin{figure}[t]
\begin{center}
    \includegraphics[width=\columnwidth]{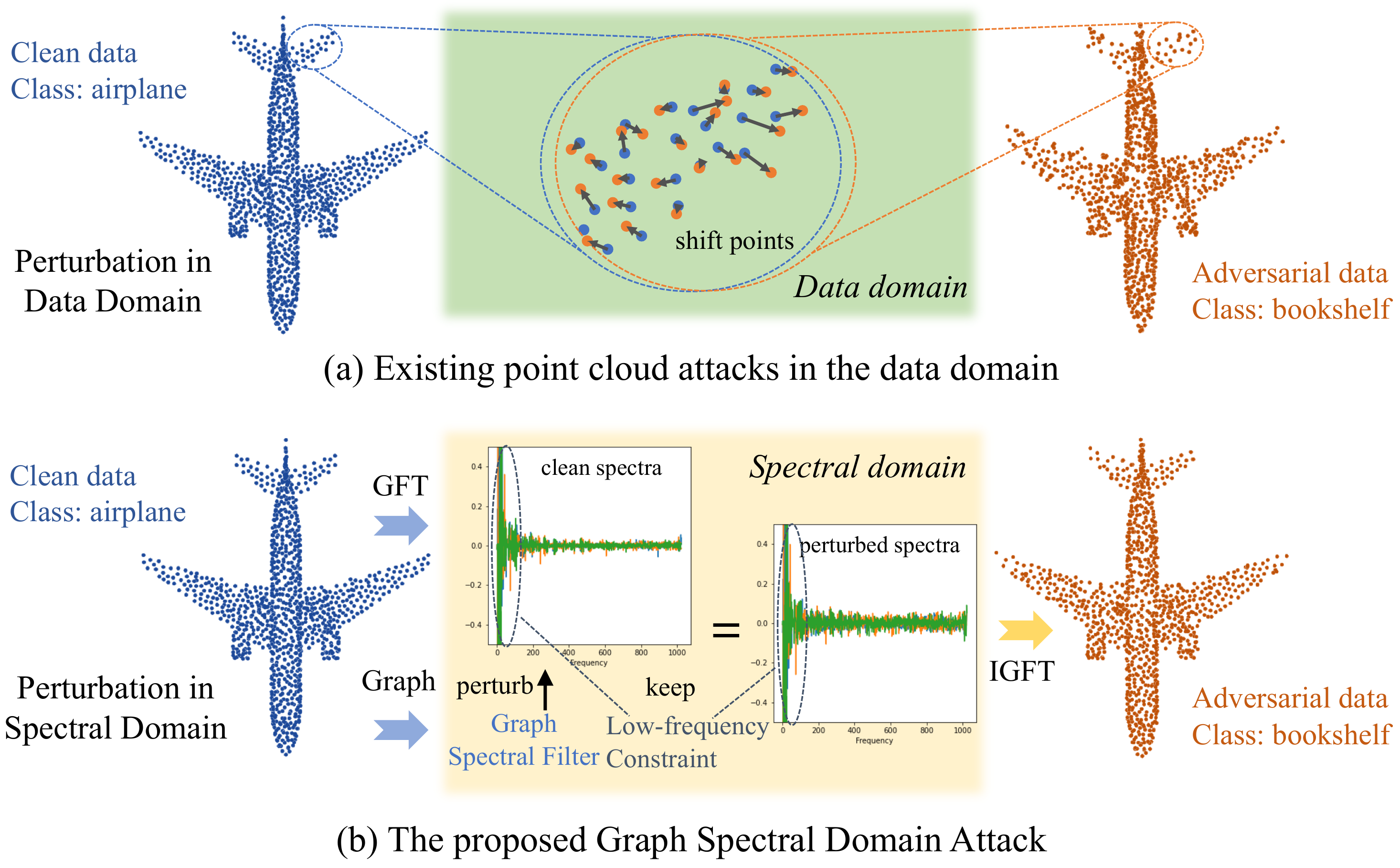}
\end{center}
\caption{(a) Existing point cloud attacks generally perturb point coordinates by shifting points in the data domain, which often fail to preserve the geometry contexts. (b) In contrast, we explore perturbations in the graph spectral domain, leading to more imperceptible and effective adversarial examples. 
When attacked with appropriate graph spectral filtering, spectra of the point cloud keeps similar distribution with the clean one, thus preserving geometric structures.
}
\label{fig:teaser}
\end{figure}

To this end, we delve into how to effectively analyze and preserve the geometry information of benign 3D objects in this paper, and propose a novel point-cloud attack method from a new perspective---Graph Spectral Domain Attack (GSDA++). 
Differently from previous methods attacking point clouds in the {\it data space}, 
\red{we propose to add perturbations in the spectral domain. This takes inspiration from the spectral-domain characteristics of point clouds, which explicitly indicate certain structural variations over
different frequency bands \cite{hu2021overview}. Specifically, the low-frequency band represents the prominent basic geometric structure of the 3D object, while the high-frequency band represents fine-grained details and possibly noise. By preserving the geometry-aware spectral information, our method is able to generate structure-preserved and thus more imperceptible adversarial samples.
This also explicitly alleviates outliers and uneven distributions.  
Therefore,} our objective is to exploit the elegant characterization of geometric structures in the spectral domain of the {\it graph representation} and thereby perturb the graph transform coefficients, which explicitly varies certain geometric structures.
That is, each graph frequency represents a certain structural variation of the point cloud, indicating the corresponding global or local geometry contexts. 
Therefore, it is crucial to investigate the frequency components and their correlations with geometry information in point clouds.
Once an appropriate frequency band is perturbed, the corresponding structural changes reflected in the data domain could be small and even imperceptible to humans. 
Hence, we first provide an in-depth graph-spectral analysis of point clouds, which shows that the rough shape of point clouds is represented by low-frequency components, while the fine details or noise of objects are encoded in high-frequency components in general.
With a trivial perturbation in this spectral domain, the point cloud could retain the original rough shape with similar local details, as shown in Figure~\ref{fig:teaser}(b).
Also, we find that the spectral characteristics of point clouds represent higher-level and global information compared to point-to-point relations captured by the distance metrics in previous work. Such a spectral representation encodes more abstract and essential contexts for recognizing 3D objects. 

Based on the above analysis, we implement the proposed GSDA++ attack with the following trainable process.
Firstly, 
we represent a point cloud on a graph naturally and adaptively, where each point is treated as a vertex and connected to its $K$ nearest neighbors, and the coordinates of each point serve as the graph signal. 
Then we compute the graph Laplacian matrix \cite{shuman2013emerging} that encodes the edge connectivity and vertex degree of the graph, whose eigenvectors form the basis of the GFT. 
Due to the compact representation of point clouds in the GFT domain \cite{hu2021overview}, we transformed the coordinate signal of point clouds into the spectral domain via the GFT, leading to transform coefficients corresponding to each spectral component.
Next, we develop a learnable Laplacian-guided spectral perturbation approach to perturb the spectral domain with adversarial noise. 
Considering that the low-frequency component of an object contains basic information while the high-frequency components represent trivial details and noise, we further propose a low-frequency constraint to limit the perturbations within imperceptible high-frequency components.
Finally, we transform the perturbed GFT coefficients back to the data domain via the inverse GFT (IGFT) to produce the adversarial point cloud.
We iteratively optimize the adversarial loss function and perform back-propagation to retrieve the gradient in the spectral domain for generating and updating the desired spectrum-aware perturbations.

Besides, existing 3D defense strategies are often insensitive to the geometry-aware frequency characteristics, as they usually reconstruct the geometry of adversarial point clouds in the data space by directly shifting point coordinates to preserve the original 3D shape, which thus fails to capture the spectral noise generated by the proposed GSDA++. 
To promote the development of 3D defense, we further propose a simple yet effective method to defend such spectral noise. 
In particular, considering that the major geometric structure of a point cloud is retained in the low-frequency components, we retrain the 3D models to focus more on the low-frequency components of point clouds for recognizing their classes. 
Specifically, we first filter out the high-frequency components of the training data and reconstruct them only with the low-frequency components. 
Then, we combine the reconstructed point clouds with their benign ones as a new training pool and feed them into the 3D models for training. 
During the inference, we only tested adversarial point clouds with their low-frequency components to capture the prominent object-aware structure information.

This work extends our previous GSDA \cite{Hu2022Exploring} in the following four aspects.
First, instead of directly adding noise to the GFT coefficients as in \cite{Hu2022Exploring}, we propose to generate spectral perturbations via a learnable graph spectral filter, which fits a desirable spectral distribution that approximates the original spectral characteristics reflecting the 3D geometry, thus preserving the corresponding geometric structure better.
Second, since the low-frequency components of an object characterize the basic shape information while the high-frequency components encode fine details and noise, we pose a low-frequency constraint to limit the perturbations within high-frequency components.
Third, we introduce a simple yet effective defense strategy to defend against such spectral attacks.
Finally, we evaluated the proposed attack on more recent victim 3D models for comparison and added more experimental results to further investigate the robustness of the proposed GSDA++.

Our main contributions are summarized as follows.
\begin{itemize}
    \item We propose a novel paradigm of spectral-domain point cloud attacks, which perturbs point clouds in the graph spectral domain to exploit the high-level spectral characterization of geometric structures. Such a spectral approach marks a significant departure from the current practice of point-cloud attacks.
    \item We provide an in-depth graph-spectral analysis of point clouds, which investigates the relations between the topology information and different frequency components and further enlightens our point-cloud attack formulation of exploring destructive perturbations on appropriate frequency components. 
    Based on this, we develop a learnable spectrum-aware perturbation approach to attack 3D models in an end-to-end manner.
    \item We develop a simple yet effective defense method to protect 3D models against such spectral domain attacks, by enforcing 3D models to focus more on the low-frequency components of point clouds. 
    \item Extensive experiments show that the proposed GSDA++ achieves 100\% of attack success rates under both targeted and untargeted settings with the least required perturbation size. We also demonstrate the imperceptibility of the GSDA++ compared to the state-of-the-arts, as well as the robustness of the GSDA++ by attacking existing defense methods. 
\end{itemize}
\section{Related Work}
\subsection{3D Point Cloud \new{Analysis}}
Deep 3D point cloud learning \cite{qi2017pointnet,qi2017pointnet++,wang2019dynamic,yang2018foldingnet,liu2019relation,rao2020global,te2018rgcnn,gao2020graphter} has received increasing attention in recent years, which has diverse applications in many fields, such as 3D object classification \cite{su2015multi,yu2018multi}, 3D scene segmentation \cite{graham20183d,wang2018sgpn,xu2020grid}, and 3D object detection in autonomous driving \cite{chen2017multi,yang2019learning}. 
Among them, 3D object classification is the most fundamental task, which learns representative information including both the local details and global context of point clouds for recognizing the object shape.
Early works attempt to classify point clouds by adapting deep learning models in the 2D space \cite{su2015multi,yu2018multi}, which max-pool multi-view object features into a global representation. However, these works suffer from high computation costs due to the usage of many 2D convolution layers. 
In order to directly extract the 3D structure and address the unorderness problem of point clouds, 
some works \cite{li2018pointcnn} follow a two-step framework that first transforms the input points into a latent and potentially canonical order and then applies the typical convolutional operator on them to learn the 3D features.
Different from the above methods,
the pioneering DeepSets \cite{zaheer2017deep} and PointNet \cite{qi2017pointnet} methods propose to achieve end-to-end learning on point cloud classification by formulating a general specification for point cloud learning.
PointNet++ \cite{qi2017pointnet++} and other extensive works \cite{duan2019structural,liu2019densepoint,yang2019modeling} are built upon PointNet to further capture the fine local structural information from the neighborhood of each point.
Recently, some works have focused on designing special convolutions in the 3D domain \cite{li2018pointcnn,atzmon2018point,thomas2019kpconv,liu2019relation} or developing graph neural networks \cite{simonovsky2017dynamic,shen2018mining,wang2019dynamic,xu2020grid,gao2020graphter,te2018rgcnn,du2021self} to improve point cloud learning.
There are also two latest models---PointTrans. \cite{zhao2021point} and PointMLP \cite{ma2022rethinking} that are recently developed based on the transformer and MLP architecture, respectively.
In this paper, we focus on PointNet \cite{qi2017pointnet}, PointNet++ \cite{qi2017pointnet++}, DGCNN \cite{wang2019dynamic}, PointTrans. \cite{zhao2021point}, PointMLP \cite{ma2022rethinking}, since these 3D models are extensively deployed in practical 3D applications.
\new{Besides, to handle the point cloud input, previous works \cite{yin2021graph} utilize a fully-connected graph structure to design message-passing networks in the data domain for encoding the features. In comparison, our work utilizes the graph Fourier transform to capture the geometric characteristics of point clouds in the spectral domain.}

\subsection{Adversarial Attack on 3D Point Clouds}
Deep neural networks are vulnerable to adversarial examples, which has been extensively explored in the 2D image domain \cite{moosavi2016deepfool,moosavi2017universal,mustafa2020deeply}.
Recently, many works \cite{xiang2019generating,wicker2019robustness,zhang2019adversarial,zheng2019pointcloud,tsai2020robust,zhao2020isometry,zhou2020lg,hamdi2020advpc,liu2023robust,tao20233dhacker} have been adapted to 2D adversarial attacks in the 3D vision community, which can be mainly divided into two categories:
1) point-addition/dropping attack:
Xiang \textit{et al.} \cite{xiang2019generating} proposed point generation attacks by adding a limited number of synthesized points/clusters/objects to a point cloud, and showed its effectiveness in attacking the PointNet model \cite{qi2017pointnet}. 
Recently, more works
\cite{zhang2019adversarial,wicker2019robustness,zheng2019pointcloud} utilize gradient-guided attack methods to identify critical points in point clouds for modification, addition, and deletion.
Their goal is to add or remove key points that can be identified by calculating the label-dependent importance score referring to the calculated gradient.
2) point perturbation attack:
Previous point-wise perturbation attacks \cite{wen2020geometry,tsai2020robust} learn to perturb xyz coordinates of each point by adopting the C\&W framework \cite{carlini2017towards} based on the Chamfer and Hausdorff distances with additional consideration of the benign distribution of points.
The subsequent works \cite{hamdi2020advpc,liu2019extending,ma2020efficient,zhang2019defense} further applied the iterative gradient method to achieve more fine-grained adversarial perturbation. 
Liu \et \cite{liu2021imperceptible} proposed to restrict point-wise perturbations along the normal direction within a strictly bounded width for preserving the geometric smoothness.
However, the above attack methods are all implemented in the {\it data space}, and their generated adversarial point clouds often result in outliers or uneven distributions, failing to preserve the original geometric information of the benign 3D object.
\red{Besides, while some previous data-domain attackers modify only a few points, these methods \cite{zhao2020nudge,zhang2019adversarial,xiang2019generating,sun2021local} need to greedy-search critical points or additionally utilize other tools to define critical points for point adding/deleting/perturbing, which are time-consuming. Their generated samples also indicate that such kind of local attack often easily results in outliers.}
In this paper, we propose to attack the 3D point cloud in a novel {\it graph spectral domain}, where the energy of each frequency band indicates a certain geometry structure of the 3D object. Once a satisfied perturbation is added on appropriate positions among spectral frequencies without disturbing the prominent characteristics, its reconstructed adversarial sample in the data domain is able to preserve the most geometry information of the benign point cloud.

\subsection{Existing Defenses on 3D Point Clouds}
\new{Most defense methods \cite{mustafa2020deeply,liao2018defense,jia2019comdefend} focus on investigating the robustness of regularly-sampled image data. In comparison, this paper focuses on the attack and defense of irregularly sampled point cloud data.}
With the rapid growth of adversarial 3D attacks, many defense methods \cite{liu2019extending,zhou2019dup,zhang2019adversarial,dong2020self,wu2020if,liu2021pointguard} proposed to defend against adversarial 3D data.
Among them, simple random sampling (SRS) \cite{zhang2019adversarial} is a widely used strategy, which first downsamples the point cloud into a smaller sub-cloud and then predicts its object class as the final result of the whole point cloud.
Zhou \et \cite{zhou2019dup} proposed both statistical outlier removal (SOR) and Denoiser and UPsampler Network (DUP-Net) defenses. SOR aims to predict the object class by removing the outlier points, while DUP-Net further develops an upsampling model to restore the removed points before feeding them into the 3D models.
Dong \et \cite{dong2020self} predicted the gather vector to distinguish the center direction of clean and adversarial local characteristics. Then it filters out adversarial local features and only aggregates clean ones for classification.
Since almost all attacks shift or modify points to add noise, their generated adversarial samples are often outliers and have uneven distribution.
Therefore, Wu \et \cite{wu2020if} proposed implicit function defense (IF-Defense) to reconstruct the adversarial point cloud by regularizing noisy points with both geometry-aware and distribution-aware loss functions.
Furthermore, Liu \et \cite{liu2021pointguard} applied certified defense to the 3D models to build the robust 3D classification model.
In this work, we evaluate the adversarial robustness of our proposed attack by testing adversarial examples on the above defense methods. 

\subsection{Spectral Methods for 3D Point Clouds}
There already exist methodologies that exploit spectral information to understand point clouds.
For example, some 3D denoising methods \cite{rosman2013patch,zhang2020hypergraph} transform the input point cloud into the graph spectral domain, where the rough shape of a point cloud is encoded into low-frequency components. The noisy shape can thus be reconstructed by the spectral filter.
Specifically, Rosman \textit{et al.} proposed spectral point cloud denoising based on the non-local framework \cite{rosman2013patch}. 
They group similar surface patches into a collaborative patch and perform shrinkage in the GFT domain by a low-pass filter, which leads to the denoising of the 3D shape. 
Zhang \textit{et al.} proposed a tensor-based method to estimate hypergraph spectral components and frequency coefficients of point clouds, which can be used to denoise 3D shapes \cite{zhang2020hypergraph}. 
Furthermore, other applications \cite{chen2017fast,ramasinghe2020spectral} also represent the fine details of point clouds through transformed high-frequency components, and use them to detect contours or process redundant information. 
In particular, Chen \textit{et al.} proposed a high-pass filtering-based resampling method to highlight contours for large-scale point cloud visualization and extract key points for accurate 3D registration \cite{chen2017fast}. 
Ramasinghe \textit{et al.} proposed Spectral-GANs to generate high-resolution 3D point clouds, which takes advantage of spectral representations for compact representation \cite{ramasinghe2020spectral}. 
In this paper, we leverage graph signal processing theory to the 3D attack task for analyzing the geometry information and perturbing learnable appropriate frequency bands of the point cloud.
\section{Graph Spectral Analysis for 3D Point Clouds}
\label{sec:analysis}
In this section, we provide in-depth graph spectral analysis for point clouds, which lays the foundation for the proposed graph spectral domain attack in Sec.~\ref{sec:attack}. 
We first provide a review of relevant concepts and tools in graph signal processing in Sec.~\ref{subsec:revisit}. Then, we elaborate on point cloud representation in the graph spectral domain and analyze the spectral characteristics of point clouds in Sec.~\ref{subsec:analysis}.

\subsection{Revisit Graph Signal Processing}
\label{subsec:revisit}
\subsubsection{Graph Variation Operators}
Formally, we denote the graph signal $\vx \in \mathbb{R}^{n}$ over a graph $ \gG=\{\gV,\gE, \mA\} $, where $ \gV $ is a vertex set of cardinality $|\gV|=n$, $ \gE $ denotes the edge set connecting the vertices, and $\mA$ is the adjacency matrix. 
Each entry $a_{i,j}$ in $\mA$ represents the weight of the edge between the vertices $i$ and $j$, capturing the similarity between these two vertices. 

Among variation operators in graph signal processing \cite{yin2021graph}, considering that point clouds are piecewise-smooth, we focus on the combinatorial graph Laplacian matrix \cite{shuman2013emerging} defined as $\mL:=\mD-\mA $, where $ \mD $ is a diagonal matrix with each entry $ d_{i,i} = \sum_{j=1}^n a_{i,j} $ denoting the degree of each node.  
Given real and non-negative edge weights in an undirected graph, $\mL$ is real, symmetric, and positive semi-definite \cite{chung1997spectral}. 
Hence, it admits an eigen-decomposition $\mL = \mU \mLambda \mU^{\top}$, where $\mU=[\vu_1,...,\vu_n]$ is an orthonormal matrix containing the eigenvectors $\vu_i$, and $\mLambda = \mathrm{diag}(\lambda_1,...,\lambda_n)$ consists of the eigenvalues $ \{\lambda_1 = 0 \leq \lambda_2 \leq... \leq \lambda_n\}$. 
In graph signal processing theory, the above eigenvalues are referred to as the {\it graph frequency/spectrum}, where a smaller eigenvalue corresponds to a lower graph frequency.

\begin{figure*}[t!]
\begin{center}
    \includegraphics[width=\textwidth]{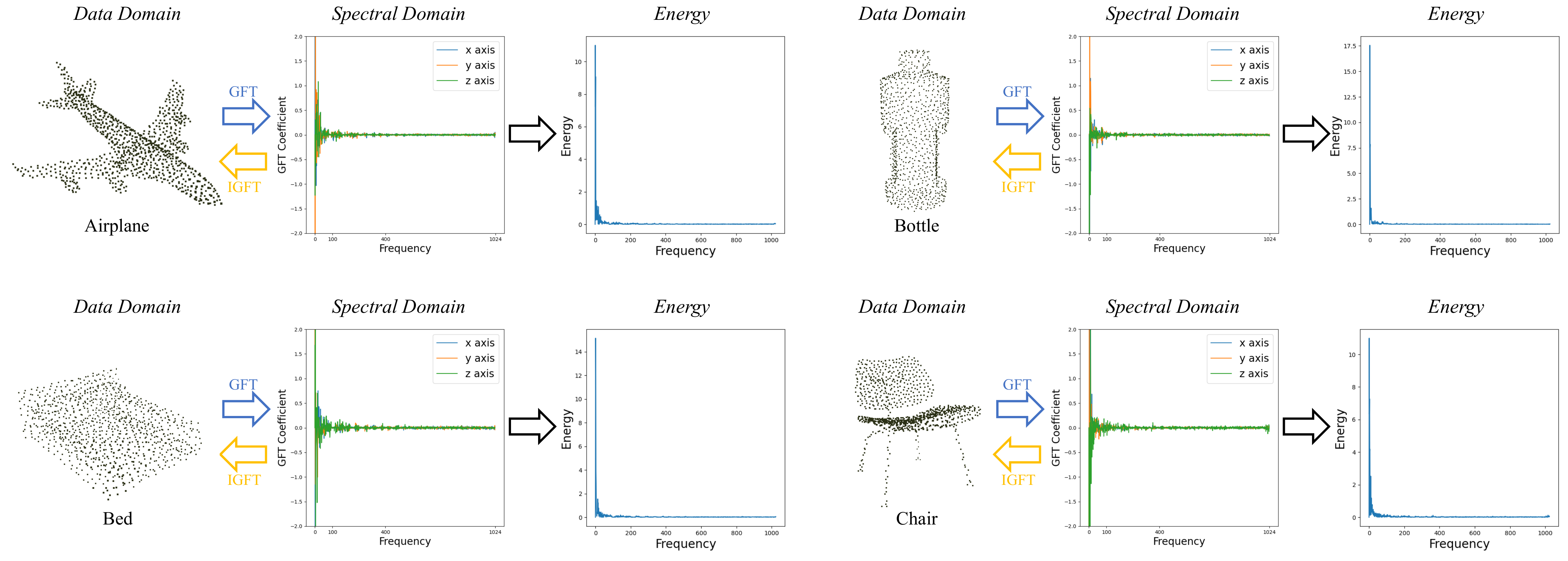}
\end{center}
\caption{Investigation on the graph spectral distribution of the 3D objects. Specifically, we transform each point cloud onto the graph spectral domain and plot their corresponding GFT coefficients and energy \new{(\textit{i.e}, the squared sum of transform coefficients)} in this figure. \new{We observe that each point cloud has larger amplitudes in lower-frequency components and much smaller amplitudes in higher-frequency components. Also, the energy mainly concentrates on the low-frequency components. This indicates that most object information of a point cloud is encoded in low-frequency bands.}}
\label{fig:energy}
\end{figure*}

\subsubsection{Graph Fourier Transform}
For any graph signal $\vx \in \mathbb{R}^{n}$ residing on the vertices of $\mathcal G$, its Graph Fourier Transform (GFT) coefficient vector $\hat{\vx}  \in \mathbb{R}^{n}$ is defined as \cite{hammond2011wavelets}:
\begin{equation}
    \hat{\vx}  = \phi_{\text{GFT}}(\vx) = \mU^{\top} \vx. 
\end{equation}
The inverse GFT (IGFT) is defined as follows:
\begin{equation}
    \vx = \phi_{\text{IGFT}}(\hat{\vx}) = \mU \hat{\vx}. 
\end{equation}
If a graph is appropriately constructed on the given input data, it is able to capture the signal structure well, and thus the GFT will lead to a compact representation of the graph signal in the spectral domain \cite{hu2014multiresolution,hu2015intra,xu2020predictive}. 
Since $\mU$ is an orthonormal matrix, both GFT and IGFT operations are lossless.

\subsubsection{Graph Spectral Filtering}
Let $h(\lambda_i) \ (i=1,2,...,n)$ denote the frequency response of a graph spectral filtering, then the filtering takes the form
\begin{equation}
\label{eq:3}
    \vx' = \mU 
    \begin{bmatrix}
    h(\lambda_1) & & \\
    & \ddots & \\
    & & h(\lambda_n) \\
    \end{bmatrix}
    \mU^{\top} \vx,
\end{equation}
where the filter first transforms the data $\vx$ onto the
GFT domain $\mU^{\top} \vx$, then performs filtering on each eigenvalue (\ie, the spectrum of the graph), and finally projects back to the spatial domain via the inverse GFT to acquire the filtered output $\vx'$.

For instance, an intuitive realization of low-pass graph spectral filtering is to completely eliminate all graph frequencies above a given bandwidth $b$ while keeping those below unchanged:
\begin{equation}
    h(\lambda_i) = 
    \begin{cases}
    0,& i > b,\\
    1,& i \leq b.
    \end{cases}
\end{equation}

\red{
Another choice to achieve graph filtering is to employ a Haar-like low-pass graph filter as:
\begin{equation}
    h(\lambda_i) = 1 - \frac{\lambda_i}{\lambda_{\text{max}}},
\end{equation}
where $\lambda_{\text{max}} = \lambda_n$ is the maximum eigenvalue for normalization, since $\lambda_1=0 \leq \lambda_2 \leq ... \leq \lambda_n$. Therefore, we have $h(\lambda_{i-1}) \geq h(\lambda_i)$. As smaller eigenvalues correspond to lower graph frequencies, lower-frequency components are preserved while higher-frequency components are weakened.}

\red{As for more complicated graph spectral filtering with desired distribution, the general design is to devise a desirable spectral distribution and then use graph filter coefficients to fit this distribution. For example, an $L$-length graph filter is in the form of a diagonal matrix:
\begin{equation}
    h(\Lambda) = \begin{bmatrix}
    \sum_{k=0}^{L-1} h_k \lambda_1^k & & \\
    & \ddots & \\
    & & \sum_{k=0}^{L-1} h_k\lambda_n^k \\
    \end{bmatrix},
\end{equation}
where $\Lambda$ is a diagonal matrix containing eigenvalues of the graph Laplacian $\bm{L}$, and $h_k$ is the $k$-th filter coefficient. If the desirable response of the $i$-th graph frequency is $c_i$, we let:
\begin{equation}
    h(\lambda_i) = \sum_{k=0}^{L-1} h_k\lambda_i^k = c_i.
\end{equation}
We solve a set of linear equations to obtain the graph filter coefficients $h_k$'s.
}

\subsection{Spectral-domain Point Cloud Analysis}
\label{subsec:analysis}
\subsubsection{Point Cloud Representation in Graph Spectral Domain}
Signals can be compactly represented in the spectral domain, provided that the transformation basis characterizes the principal components of the signals. 
Unlike images supported on regular grids, point clouds reside in irregular domains with no ordering of points, which hinders the deployment of traditional transforms such as the DCT. 
Though we may quantize point clouds onto regular voxel grids or project onto a set of depth images from multiple viewpoints, this would inevitably introduce quantization loss.
In contrast, graphs provide {\it accurate}, {\it structure adaptive}, and {\it compact} representation for point clouds \cite{hu2021overview}. 
Hence, it is advantageous to represent point clouds on the GFT domain \cite{shen2010edge,hu2012depth,hu2014multiresolution,hu2015intra,zhang2014point,xu2019predictive}.

\begin{figure*}[t!]
\begin{center}
    \includegraphics[width=\textwidth]{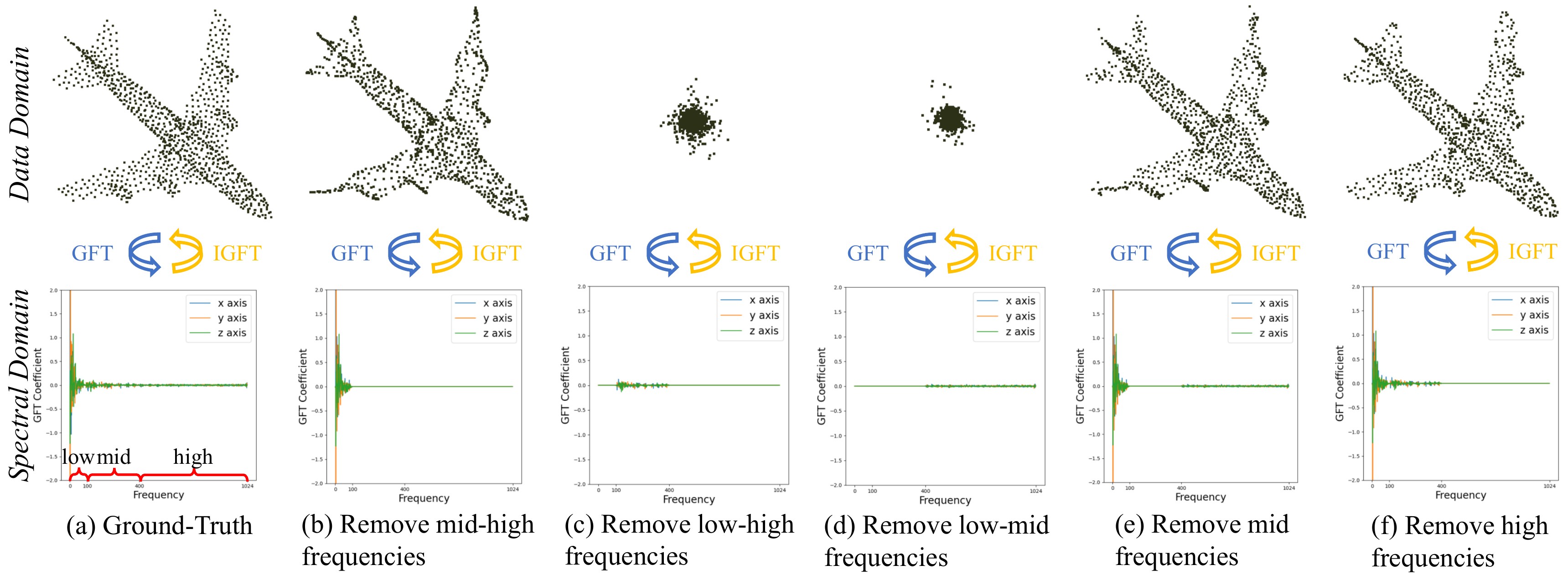}
\end{center}
\caption{Graph spectral analysis for 3D point clouds. We take an example from the object \textit{airplane} to investigate the roles of different frequency bands by removing certain frequency components in the graph spectral domain. (a) Ground-Truth; (b) Remove mid- and high-frequencies; (c) Remove low- and high-frequencies; (d) Remove low- and mid-frequencies; (e) Remove mid-frequencies; (f) Remove high-frequencies.}
\label{fig:remove}
\end{figure*}

In particular, given a point cloud $\mP=\{\vp_i\}_{i=1}^n \in \mathbb{R}^{n \times 3}$, we represent it over a graph $ \gG=\{\gV,\gE, \mA\} $. Here, we construct an unweighted $K$-nearest-neighbor graph ($K$-NN graph), where each vertex is connected to its $K$ nearest neighbors in terms of the Euclidean distance with weight $1$. The coordinates of points in $\mP$ are treated as graph signals. 
Hence, we obtain its GFT coefficient vector by $\hat{\bm{P}}=\phi_{\text{GFT}}(\bm{P}) = \bm{U}^{\top} \bm{P}$ and its IGFT  $\phi_{\text{IGFT}}(\hat{\bm{P}}) = \bm{U} \hat{\bm{P}}$.

\subsubsection{Analysis in Graph Spectral Domain}
In general, when an appropriate graph is constructed that captures the geometric structure of point clouds well, the low-frequency components of the corresponding GFT mainly characterize the {\it rough shape} of point clouds, while the high-frequency components represent {\it fine details or noise} (\ie, large variations such as geometric contours) of the 3D objects. 
This is because the variation of the eigenvectors (\ie, the GFT basis) of the graph Laplacian matrix gradually increases from low frequencies to high frequencies, capturing more and more detailed structure of the 3D object. 
For example, all values in the first eigenvector are $1$, which represents the smooth surface, while values in the last eigenvector are alternately positive and negative for representing fine details.

Further, we provide an intuitive example with in-depth analysis for the understanding of the spectral characteristics, \ie, how each frequency band contributes to the geometric structure of a point cloud in the data domain.
Specifically, we randomly take some point clouds from the ModelNet40 dataset \cite{wu20153d}, and sample each of them into $1024$ points as an example point cloud $\bm{P}$.
We construct a $K$-NN graph ($K=10$) in the point cloud and perform the GFT on the three coordinate signals of each point in $\bm{P}$. 
Examples of the resulting transform coefficient vectors $\phi_{\text{GFT}}(\bm{P})$ in the spectral domain are presented in Figure~\ref{fig:energy}.
We see that, $\phi_{\text{GFT}}(\bm{P})$ has larger amplitudes in lower-frequency components and much smaller amplitudes in higher-frequency components, demonstrating that most information is concentrated in low-frequency components.
That is, a point cloud is a {\it low-pass} signal when the graph is appropriately constructed.  
Since there is no official principle for the definition of different frequency bands, we propose to divide the entire spectral domain into three bands via the distribution of energy---the squared sum of transform coefficients. 
As shown in Figure~\ref{fig:energy}, point clouds have almost 75\% of energy within the lowest 100 frequencies and almost 90\% of energy within the lowest 400 frequencies. 
Based on this observation, we set three frequency bands for point clouds in the ModelNet40 dataset: the low-frequency band (frequency $[0,100)$), the mid-frequency band (frequency $[100,400)$), and the high-frequency band (frequency $[400,1024)$).

\begin{figure}[t!]
\begin{center}
    \includegraphics[width=0.5\textwidth]{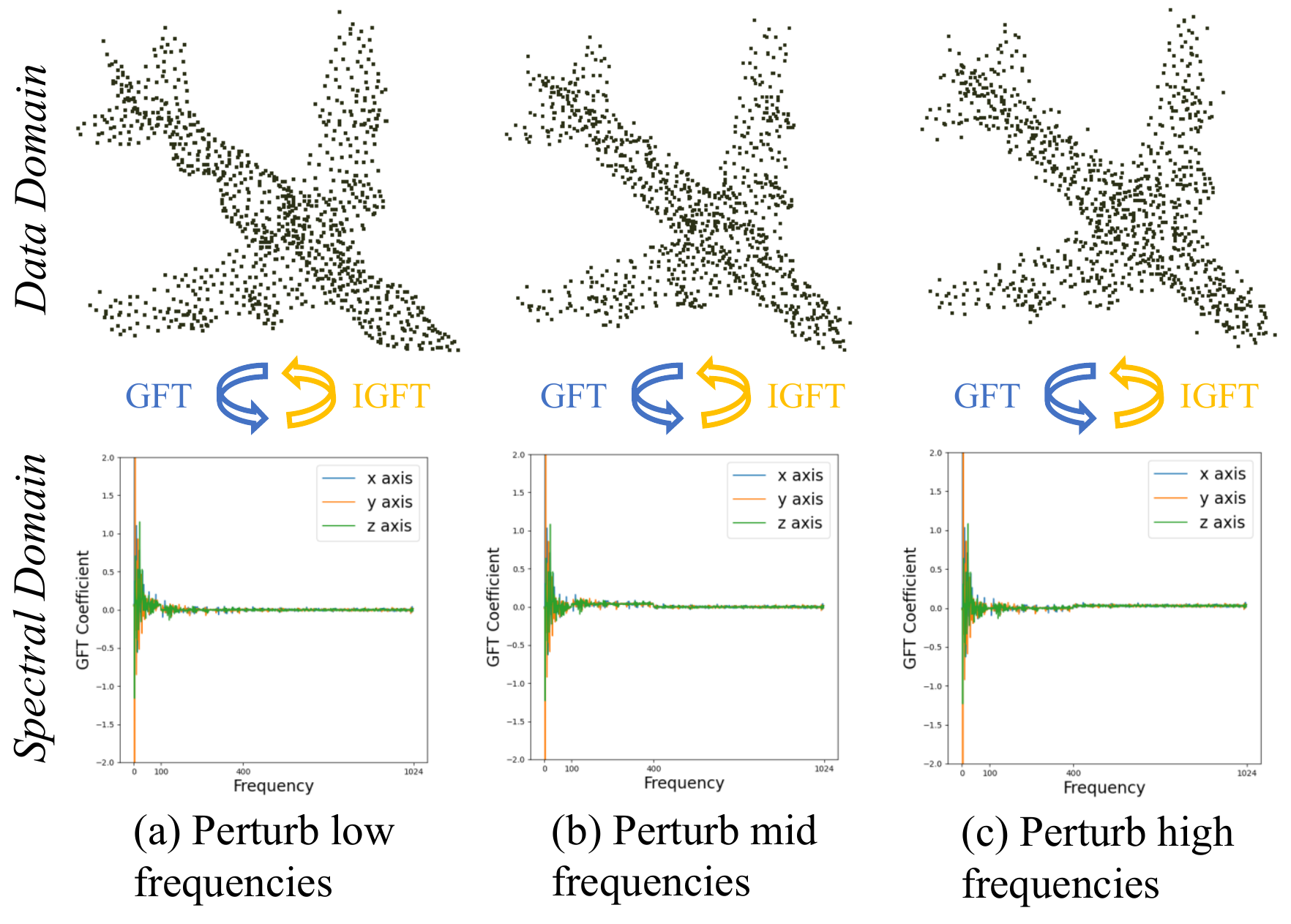}
\end{center}
\caption{Investigation of geometric changes in the data domain by different frequency-sensitive perturbations. (a) Perturb the low-frequency band; (b) Perturb the mid-frequency band; (c) Perturb the high-frequency band.}
\label{fig:perturb}
\end{figure}

We study the influence of each frequency band on the geometric structure by setting certain GFT coefficients to zero via graph spectral filtering.
As shown in Figure~\ref{fig:remove} (b), when the GFT coefficients in the mid- and high-frequency bands are set to zero, the point cloud reconstructed with only the low-frequency band exhibits the rough shape of the original object without fine details such as in the airfoils. 
Figure~\ref{fig:remove} (c) and (d) show the point cloud reconstructed with only the mid- or high-frequency band, respectively, further indicating the importance of the low-frequency band for constructing the object shape.
By adding more information from the mid- or high-frequency bands to (b), the reconstructed point cloud has richer local contexts, as shown in Figure~\ref{fig:remove}(e)(f), but still lacks fine-grained details, such as the engines of the airplane. 

In summary, each frequency band represents different aspects of the geometry of a point cloud. In particular, the low-frequency band represents the prominent geometric structure of the 3D object, the mid-frequency band contributes to some shape details in each part, while the high-frequency band represents fine details and noise.

\subsubsection{Insight for Point Cloud Attacks}
Enlightened by the above analysis, the perturbations of different frequency components would result in various geometric changes in the data space. 
Here is the key question: {\it what are the results of attacking different frequency bands?} 
As demonstrated in Figure~\ref{fig:perturb}, we investigate into this by separately perturbing each frequency band with a certain perturbation size, \ie, we equally perturb each frequency in a specific band by adding a certain amount of noise.
We observe that, attacking low-frequency components introduces deformation in the rough shape but maintains the smoothness of the surface. 
Perturbing mid-frequency or high-frequency components loses local details and induces noise and outliers, though the silhouette of the shape is preserved to some degree thanks to the clean lower-frequency components.

\begin{figure*}[t!]
    \centering
    \includegraphics[width=\textwidth]{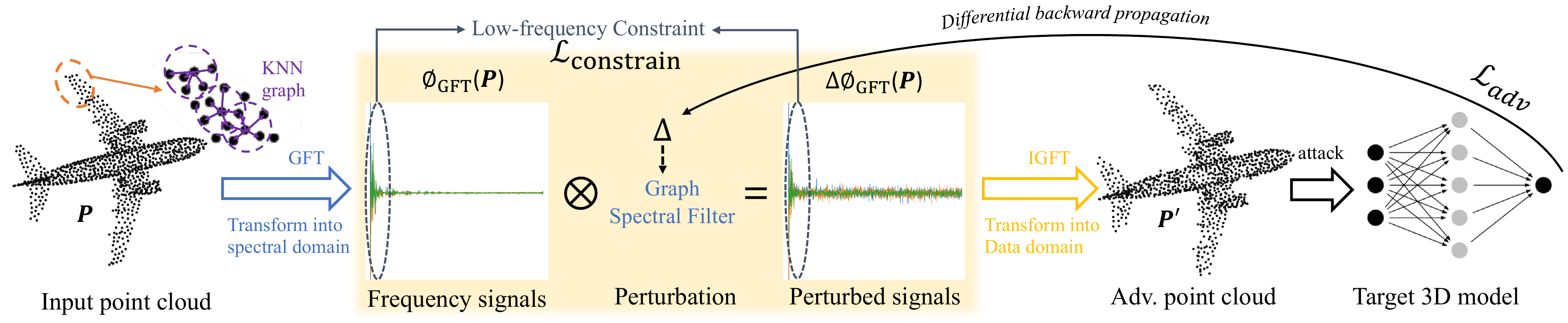}
    \caption{{The overall pipeline of the proposed GSDA++.} Given a clean point cloud, we first construct a $K$-NN graph and transform the point cloud onto the spectral domain via the GFT. Then, we perturb the GFT coefficients in a learnable manner with a specifically designed graph spectral filter. Subsequently, we transform the perturbed spectral signals back to the data domain via the IGFT. Finally, we take the reconstructed point cloud as the adversarial example and feed it into the target 3D model for attack.}
    \label{fig:pipeline}
\end{figure*}

Inspired by the above properties of point clouds in the graph spectral domain, we summarize the key insights for developing effective spectral domain attacks:
\begin{itemize}
    \item Each frequency band represents the geometric structure of a point cloud from different perspectives: the low-frequency components represent the basic shape while the mid-/high-frequency components encode find-grained details. 
    The perturbation of certain frequency band may result in the corresponding distortion in the data domain.
    \item Although a large perturbation size can ensure a high success attack rate, it may severely change the spectral characteristics, thus leading to perceptible deformations in the data domain.
    \item A relatively small perturbation in the low-frequency components is able to preserve the prominent shape of an object.
\end{itemize}

Based on the above insights, a desirable spectral domain attack for point clouds need to not only perform the perturbation among appropriate frequencies for striking a balance, but also restrict the perturbations especially in the low-frequency band for preserving the original spectral characteristics and thus the shape in the data domain. 

\section{Graph Spectral Domain Attack}
\label{sec:attack}
Leveraging on the insights in Section~\ref{sec:analysis}, we propose a novel 3D attack that selectively perturbs a point cloud in the graph spectral domain to keep the prominent geometric information.
Specifically, we learn perturbations over the graph frequencies (\ie, eigenvalues of the graph Laplacian matrix) for a learnable distribution in order to transform the latent feature into another class-specific one.
To preserve the basic geometric structure of the original 3D object, we further develop a low-frequency constraint to limit the perturbations within imperceptible high-frequency components.
In the following, we first elaborate on the formulation of the proposed point cloud attack, and then provide the detailed algorithm.

\subsection{The Formulation of Our Attack Method}
Given a clean point cloud $\bm{P}=\{\bm{p}_i\}_{i=1}^n \in \mathbb{R}^{n \times 3}$ where each point $\bm{p}_i \in \mathbb{R}^3$ is a vector that contains the coordinates (x, y, z), a well-trained point cloud classifier $f(\cdot)$ can predict its accurate label $y=f(\bm{P})\in \mathbb{Y},\mathbb{Y}=\{1,2,3,...,c\}$ that best represents its object class, where $c$ is the number of classes.
The goal of point cloud attacks on classification is to deform the point cloud $\bm{P}$ into an adversarial one $\bm{P}'$, so that $f(\bm{P}')=y'$ (targeted attack) or $f(\bm{P}')\neq y$ (untargeted attack), where $y' \in \mathbb{Y}$ but $y' \neq y$.


We propose a novel graph spectral domain attack (GSDA++) that aims to learn destructive yet imperceptible perturbations in the spectral domain to generate adversarial point clouds. 
In particular, the objective is to learn perturbations in the spectral domain that minimize the adversarial loss for preserving the geometric characteristics of the 3D object, under the distance constraint between the GFT coefficients before and after the attack. 
The attack is realized by graph spectral filtering in the GFT domain. 
Formally, we formulate the proposed GSDA++ as the following optimization problem:
\begin{equation}
\label{eq:objective}
\begin{aligned}
    & \min_{\bm{\Delta}} \mathcal{L}_{adv}(\bm{P}',\bm{P},y), ~~\text{s.t.}~~ ||\phi_{\text{GFT}}(\bm{P}')-\phi_{\text{GFT}}(\bm{P})||_p < \epsilon, \\
    & \text{where} \ \bm{P}' = \phi_{\text{IGFT}}
    \bm{\Delta}
    (\phi_{\text{GFT}}(\bm{P})), \\
    & \bm{\Delta} = \begin{bmatrix}
    \bm{\Delta}_{w,1}\cdot \sum_{l=0}^{L-1} \bm{\Delta}_{h,l} \lambda_1^l & & \\
    & \ddots & \\
    & & \bm{\Delta}_{w,n} \cdot \sum_{l=0}^{L-1} \bm{\Delta}_{h,l}\lambda_n^l \\
    \end{bmatrix},
\end{aligned}
\end{equation}
where $\mathcal{L}_{adv}(\bm{P}',\bm{P},y)$ is the adversarial loss and $\bm{\Delta}$ is the learnable perturbation in the spectral domain. 
In the imposed constraint, $\epsilon$ is a threshold that aims to restrict the size of the perturbation in the spectral domain, which preserves the original spectral characteristics so that the resultant adversarial point cloud $\bm{P}'$ is visually indistinguishable from its clean version $\bm{P}$. 
The distance is measured in the $l_p$ norm, where we adopt the $l_2$ norm.
To achieve the perturbation in the spectral domain, we perform filtering on each GFT coefficient as in Eq.~(\ref{eq:3}), where the graph spectral filter is a polynomial function of the eigenvalues so as to fit a desirable spectral distribution. 
Specifically, $\{\bm{\Delta}_{w,i}\}_{i=1}^n$ highlights the contributed frequency components. 

To back-propagate the gradient in a desired direction for optimizing the perturbation learning, we define our adversarial loss $\mathcal{L}_{adv}(\bm{P}',\bm{P},y)$ as follows:
\begin{equation}
\label{eq:adv_loss}
\begin{aligned}
    \mathcal{L}_{adv}(\bm{P}',\bm{P},y) = & \mathcal{L}_{class}(\bm{P}',y) + \beta_1 \cdot \mathcal{L}_{reg}(\bm{P}',\bm{P}) \\
    & + \beta_2 \cdot \mathcal{L}_{constrain}(\widetilde{\bm{P}'},\widetilde{\bm{P}}),
\end{aligned}
\end{equation}
where $\mathcal{L}_{class}(\bm{P}',y)$ promotes the misclassification of the point cloud $\bm{P}'$. $\mathcal{L}_{reg}(\bm{P}',\bm{P})$ is a regularization term that minimizes the distance between $\bm{P}'$ and $\bm{P}$ to guide the perturbation at appropriate frequencies. $\mathcal{L}_{constrain}$ is the proposed low-frequency constraint to preserve the shape of the 3D object and limit the perceptible noise, which will be discussed in Section~\ref{sec:low} in detail. $\widetilde{\bm{P}'},\widetilde{\bm{P}}$ are the reconstructed point cloud based on only the low-frequency components of $\bm{P}',\bm{P}$, respectively. 
$\beta_1,\beta_2$ are penalty parameters. 

Specifically, $\mathcal{L}_{class}(\bm{P}',y)$ is formulated as a cross-entropy loss as follows:
\begin{equation}
    \mathcal{L}_{class}(\bm{P}',y) = \left\{
    \begin{aligned}
        -\log(p_{y'}(\bm{P}')), \quad &\text{for targeted attack,} \\
        \log(p_{y}(\bm{P}')), \quad &\text{for untargeted attack,}
    \end{aligned}
    \right.
\end{equation}
where $p(\cdot)$ is the softmax functioned on the output of the target model, \ie, the probability with respect to adversarial class $y'$ or clean class $y$. 
By minimizing this loss function, the proposed attack optimizes the spectral perturbation $\bm{\Delta}$ to mislead the target model $f(\cdot)$.

In addition, to strike a balance of perturbing different frequency bands, we employ both the Chamfer distance loss \cite{fan2017point} and Hausdorff distance loss \cite{huttenlocher1993comparing} for $\mathcal{L}_{reg}(\bm{P}',\bm{P})$.
This reflects the imperceptibility in the data domain for updating the frequency perturbation $\bm{\Delta}$.

\subsection{Low-Frequency Constraint of GSDA++}
\label{sec:low}
Although the perturbations are added and optimized in the graph spectral domain, there is still a risk that these perturbations may distort some spectral characteristics and lead to perceptible structural changes in local regions. 
Conventional constraints, on the other hand, may result in a random distribution of perturbations. 
Hence, we seek a new constraint to limit perturbations within imperceptible details of the original point clouds.
Since the low-frequency components mainly contribute to the rough shape of the 3D object, we impose the constraint on these components, which guides the perturbation to concentrate on the high-frequency components that represent fine details and noise.

Specifically, in order to preserve the prominent structure information of the original object, we set the high-frequency components of both the benign point cloud $\bm{P}$ and its adversarial sample $\bm{P}'$ to zero, and reconstruct the new point cloud with only their low-frequency components as follows:
\begin{equation}
\label{eq:low}
\begin{aligned}
    & \widetilde{\bm{P}} = 
    \mU 
    \begin{bmatrix}
    h(\lambda_1) & & \\
    & \ddots & \\
    & & h(\lambda_n) \\
    \end{bmatrix}
    \mU^{\top} \bm{P}, \\
    & \widetilde{\bm{P}'} = 
    \mU 
    \begin{bmatrix}
    h(\lambda_1) & & \\
    & \ddots & \\
    & & h(\lambda_n) \\
    \end{bmatrix}
    \mU^{\top} \bm{P}', \\
\end{aligned}
\end{equation}
where $h(\lambda_i)$ is a low-pass graph filter as follows:
\begin{equation}
    h(\lambda_i) = 
    \begin{cases}
    1,& i < b,\\
    0,& i \geq b.
    \end{cases}
\end{equation}
Here, $b$ is the upper bound of the low-frequency band and set to $b=400$ in the experiments.  

Hence, the proposed low-frequency constraint between the benign and adversarial point clouds is
\begin{equation}
    \mathcal{L}_{constrain}(\widetilde{\bm{P}'},\widetilde{\bm{P}}) = ||\widetilde{\bm{P}} - \widetilde{\bm{P}'} ||_2,
\end{equation}
which is enforced during the generation process of adversarial examples.

\subsection{Overall Algorithm of the Proposed GSDA++}
Based on the problem formulation, we develop an efficient and effective algorithm for the GSDA++ model.  
As shown in Figure~\ref{fig:pipeline}, the proposed GSDA++ attack is composed of three steps:
1) First, the GSDA++ transforms the clean point cloud $\bm{P}$ from the data domain to the graph spectral domain via the GFT operation $\phi_{\text{GFT}}$.
2) Then, the GSDA++ perturbs the GFT coefficients through our designed perturbation strategy as in Eq.~(\ref{eq:objective}). 
3) Finally, we convert the perturbed spectral signals back to the data domain using the IGFT operation $\phi_{\text{IGFT}}$ to construct the adversarial point cloud $\bm{P}'$. 
We optimize the adversarial loss function to iteratively update the desired perturbations added in the spectral domain. 
In the following, we elaborate on each module in order.
\vspace{4pt}

\noindent \textbf{Transform onto the Spectral Domain.}
Given a clean point cloud $\mP$, we employ the GFT to transform $\mP$ onto the graph spectral domain. 
Specifically, we first construct a $K$-NN graph on the whole point cloud and then compute the graph Laplacian matrix $\mathbf{L}$. 
Next, we perform eigen-decomposition to acquire the orthonormal eigenvector matrix $\bm{U}$, which serves as the GFT basis. 
The GFT coefficients $\phi_{\text{GFT}}(\bm{P})$ are then obtained by:
\begin{equation}
    \phi_{\text{GFT}}(\bm{P}) = \bm{U}^{\top} \bm{P},
\end{equation}
where $\phi_{\text{GFT}}(\bm{P}) \in \mathbb{R}^{n \times 3}$ corresponds to the transform coefficients of the x, y, z coordinate signals.

\vspace{4pt}

\noindent \textbf{Perturbation in the Graph Spectral Domain.}
We deploy a trainable perturbation $\bm{\Delta}$ to perturb the spectral representation of $\bm{P}$:  $\bm{\Delta}( \phi_{\text{GFT}}(\bm{P}))$.
To adaptively adjust the perturbation $\bm{\Delta}$ and improve the success rate of the proposed spectral attack, we solve the optimization problem in Eq.~\ref{eq:objective} by leveraging the gradients of the target 3D model $f(\cdot)$ (\eg, PointNet) through backward propagation.
During the adaptive learning process, we iteratively adjust the perturbation $\bm{\Delta}$ to increase the success rate of the attack.
Specifically, we update the perturbation $\bm{\Delta}$ with the gradients and learn the perturbation $\bm{\Delta}$ as:
\begin{equation}
\label{eq:update_delta}
\begin{aligned}
    &\bm{\Delta}' \leftarrow \bm{\Delta} + lr \cdot \partial_{\bm{\Delta}}(\mathcal{L}_{adv}(\bm{P}',\bm{P},y)),
\end{aligned}
\end{equation}
where $lr$ is the learning rate. 
\vspace{4pt}

\noindent \textbf{Inverse Transform onto the Data Domain.}
After obtaining the perturbed spectral representations, we apply the IGFT to convert the perturbed signals from the spectral domain back to the data domain as:
\begin{equation}
    \bm{P}' = \phi_{\text{IGFT}}\bm{\Delta} (\phi_{\text{GFT}}(\bm{P})) = \bm{U} \bm{\Delta}  (\bm{U}^{\top} \bm{P}),
\end{equation}
where $\bm{P}'$ is the crafted adversarial point cloud.

\begin{figure}[t!]
\begin{center}
    \includegraphics[width=0.5\textwidth]{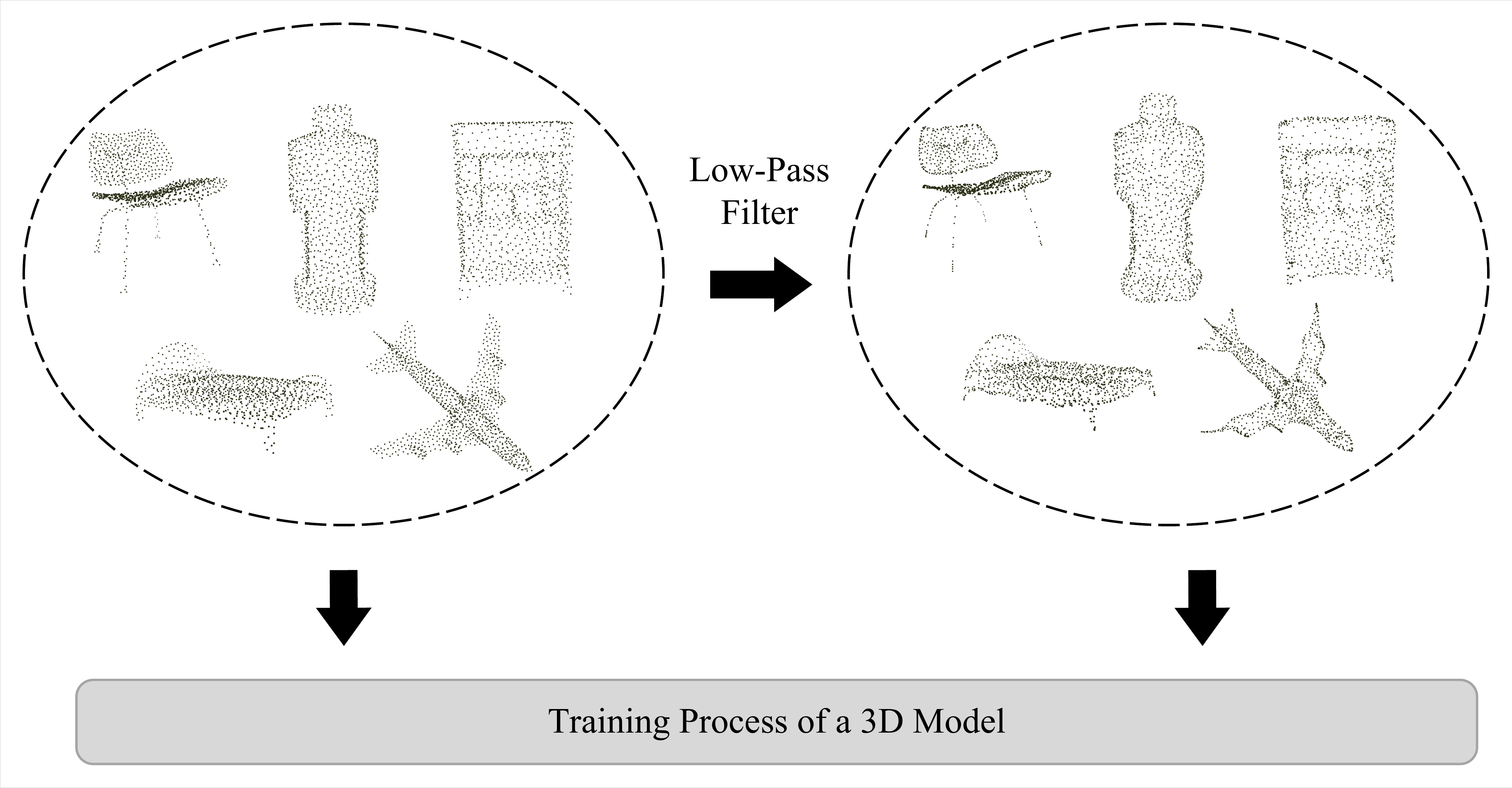}
\end{center}
\caption{Illustration of our defense strategy.}
\label{fig:defense}
\end{figure}

\subsection{Properties of the Proposed GSDA++}
Since the proposed GSDA++ takes advantage of the spectral information to implement the attack, it has several important properties. 

1) Robustness to {\em scaling}: Most previous attackers perturb point clouds by shifting the xyz coordinates of each point. Applying scaling to their adversarial samples may destroy the noise distribution and fail to achieve successful attacks. 
In contrast, our GSDA++ first builds a $K$-NN graph on each point to capture the local contexts, and then transforms the point cloud onto the spectral domain for perturbations. Simply scaling the point clouds does not change the $K$-nearest neighbors of each point, thus maintaining the same $K$-NN graph as the original one. Therefore, the spectral frequencies of the scaled point cloud and its original one are the same, which makes our GSDA++ robust to scaling. 

2) Robustness to {\em rotation}: Similar to the scaling process, rotation does not alter the $K$-NN graph. Thus, our GSDA++ is robust to 3D rotations. 
Note that, our GSDA++ design is orthogonal to the recently developed ART-Point that achieves rotation robustness through adversarial learning \cite{wang2022art}. 
There is room for further improvement by combining the two complementary strategies. 

To verify the above properties, we conduct detailed experiments in Sec.~\ref{subsec:robustness}.
\section{Defense to the Spectral Attack}
So far, we have generated high-quality adversarial point clouds in the graph spectral domain using our GSDA++ attack.
Since existing defenses on point clouds \cite{liu2019extending,zhou2019dup,zhang2019adversarial,dong2020self,wu2020if,liu2021pointguard} generally employ empirical operations (\eg, point cloud sampling, point removal, and denoising) in the data space to reconstruct the geometry of the point cloud, they are insensitive to spectral noise, and thus may be ineffective for the proposed GSDA++.
To promote the development of 3D defense against such spectral attack, we develop a simple yet effective defense method. 

Based on the analysis in Section~\ref{sec:analysis}, perturbations concentrate more on the mid- and high-frequency components of the adversarial point clouds so as to retain most of the low-frequency characteristics to keep the prominent shape information. 
Hence, it is intuitive to make the considered 3D classification model sensitive to the low-frequency components of each point cloud.
As demonstrated in Figure~\ref{fig:defense}, given the training set of point clouds, we deploy a low-pass filter to remove all mid- and high-frequency components, and reconstruct their corresponding low-pass point clouds following Eq.~(\ref{eq:low}).
We then feed a mixture of the original point cloud and its reconstructed low-pass one into the 3D model for training. 
This enforces the model training to focus more on the low-frequency information of the original data; thus, the trained model is more robust to the spectral attack. 

\section{Experiments}
\subsection{Dataset and 3D Models}
\noindent \textbf{Dataset.}
Following previous works, we adopt the point cloud benchmark ModelNet40 \cite{wu20153d} dataset in all the experiments. 
This dataset contains 12311 CAD models from 40 most common object categories in the world. 
Among them, 9843 objects are used for training and the other 2468 for testing. 
As in a previous work \cite{qi2017pointnet}, we uniformly sample $n=1,024$ points from the surface of each object and rescale them into a unit ball. 
For adversarial point-cloud attacks, we follow \cite{xiang2019generating,wen2020geometry} and randomly select 25 instances for each of 10 object categories in the ModelNet40 testing set, which can be well classified by the classifiers of interest.

\noindent \textbf{3D Models.}
We select five commonly used networks in the 3D computer vision community as victim models, \ie, PointNet \cite{qi2017pointnet}, PointNet++ \cite{qi2017pointnet++}, DGCNN \cite{wang2019dynamic}, PointTrans. \cite{zhao2021point} and  PointMLP \cite{ma2022rethinking}. 
We train them from scratch, and the test accuracy of each trained model is within 0.1\% of the best accuracy reported in their original articles. 

\subsection{Implementation Details}
\noindent \textbf{Experimental Settings.} To generate adversarial examples, we update the frequency perturbation $\bm{\Delta}$ with $500$ iterations. We use Adam optimizer \cite{kingma2014adam} to optimize the objective of our proposed GSDA++ attack in Eq.~(\ref{eq:objective}) with a fixed learning rate $lr=0.01$, and the momentum is set as $0.9$. We set the number $L$ in Eq.~(\ref{eq:objective}) as 5.
We assign $K=10$ to build a $K$-NN graph. The penalty parameters $\beta_1,\beta_2$ in Eq.~(\ref{eq:adv_loss}) is initialized as $10,1$ and adjusted by 10 runs of binary search~\cite{madry2017towards}. 
The weights of the Chamfer distance loss \cite{fan2017point} and Hausdorff distance loss \cite{huttenlocher1993comparing} in $\mathcal{L}_{reg}(\bm{P}',\bm{P})$ of Eq.~(\ref{eq:adv_loss}) are set to $5.0$ and $0.5$, respectively.
Since the targeted attack is more challenging than the untargeted attack, we focus on the targeted attack in the experiments.
All experiments are implemented on a single NVIDIA RTX 2080Ti GPU.

\noindent \textbf{Evaluation Metrics.}
To quantitatively evaluate the effectiveness of our proposed GSDA++ attack, we measure the attack success rate, which is the ratio of successfully fooling a 3D model. 
In addition, to measure the perturbation size of different attackers, we adopt five evaluation metrics: (1) Data domain: $l_2$ norm distance $\mathcal{D}_{norm}$ \cite{cortes2012l2}, Chamfer distance $\mathcal{D}_{c}$ \cite{fan2017point}, Hausdorff distance $\mathcal{D}_{h}$ \cite{huttenlocher1993comparing} and Geometric regularity $\mathcal{D}_{g}$ \cite{wen2020geometry}; (2) Spectral domain: perturbed energy $\mathcal{E}_{\Delta}=||\phi_{\text{GFT}}(\bm{P}')-\phi_{\text{GFT}}(\bm{P})||_2$. 

\begin{table}[t!]
\centering
\caption{\new{Comparison on perturbation sizes of competitive point cloud attack methods.}}
\setlength{\tabcolsep}{1.6mm}{
\begin{tabular}{ccccccc}
\hline
\multirow{2}{*}{\begin{tabular}[c]{@{}c@{}}Attack\\ Model\end{tabular}} & \multirow{2}{*}{Methods} & \multirow{2}{*}{\begin{tabular}[c]{@{}c@{}}Success\\ Rate\end{tabular}} & \multicolumn{4}{c}{Perturbation Size} \\ \cline{4-7} 
&                          &                                                                         & $\mathcal{D}_{norm}$     & $\mathcal{D}_{c}$       & $\mathcal{D}_{h}$ & $\mathcal{D}_{g}$      \\ \hline
\multirow{8}{*}{PointNet}                                               & FGSM                     & 100\%                                                                   & 0.7936      & 0.1326     & 0.1853 & 0.3901    \\
                                                                        & 3D-ADV                   & 100\%                                                                   & 0.3032      & 0.0003     & 0.0105  & 0.1772    \\
                                                                        & GeoA                     & 100\%                                                                   & 0.4385      & 0.0064     & 0.0175  & 0.0968   \\
                                                  ~ & \red{AOF} & \red{100\%} & \red{0.4142} & \red{0.0179} & \red{0.0336} & \red{0.1257} \\
~ & \red{L3A} & \red{100\%} & \red{0.3018} & \red{0.0025} & \red{0.0154} & \red{0.1361} \\
~ & \red{SI-ADV} & \red{100\%} & \red{0.2997} & \red{\textbf{0.0002}} & \red{0.0228} & \red{0.0985} 
\\
& GSDA                     & 100\%                                                                   & 0.1741      & 0.0007     & 0.0031  & 0.0817   \\
& GSDA++                     & 100\%                                                                   &     \textbf{0.1517}        & 0.0006              & \textbf{0.0028} & \textbf{0.0633}    \\ \hline
\multirow{8}{*}{PointNet++}                                             & FGSM                     & 100\%                                                                   & 0.8357      & 0.1682     & 0.2275  & 0.4143   \\
                                                                        & 3D-ADV                   & 100\%                                                                   & 0.3248      & 0.0005     & 0.0381   & 0.2034  \\
                                                                        & GeoA                     & 100\%                                                                   & 0.4772      & 0.0198     & 0.0357  & 0.1141   \\
                                                                        & \red{AOF}  & \red{100\%} & \red{0.4471} & \red{0.0294} & \red{0.0512} & \red{0.1648} \\
~ & \red{L3A} & \red{100\%} & \red{0.3283} & \red{0.0037} & \red{0.0290} & \red{0.1625} \\
~ & \red{SI-ADV} & \red{100\%} & \red{0.3213} & \red{\textbf{0.0004}} & \red{0.0342} & \red{0.1209}\\
                                           & GSDA                    & 100\%                                                                   & 0.2072      & 0.0081     & 0.0248  & 0.1075   \\                             & GSDA++                     & 100\%                                                                   &     \textbf{0.1664}        &     0.0065       & \textbf{0.0128} & \textbf{0.0986}           \\ \hline
\multirow{8}{*}{DGCNN}                                                  & FGSM                     & 100\%                                                                   & 0.8549      & 0.1890      & 0.2506   & 0.4217  \\
                                                                        & 3D-ADV                   & 100\%                                                                   & 0.3326      & \textbf{0.0005}     & 0.0475  & 0.2019    \\
                                                                        & GeoA                     & 100\%                                                                   & 0.4933      & 0.0176     & 0.0402  & 0.1174   \\
                                           & \red{AOF} & \red{100\%} & \red{0.4583} & \red{0.0265} & \red{0.0370} & \red{0.1619}\\
~ & \red{L3A} & \red{100\%} & \red{0.3306} & \red{0.0034} & \red{0.0301} & \red{0.1597} \\
~ & \red{SI-ADV} & \red{100\%} & \red{0.2884} & \red{0.0006} & \red{0.0219} & \red{0.1120}  \\
                                           & GSDA                     & 100\%                                                                   & 0.2160      & 0.0104     & 0.1401  & 0.1129  \\                             & GSDA++                      & 100\%                                                                   &     \textbf{0.1731}        & 0.0072     & \textbf{0.0135} & \textbf{0.0960}    \\ \hline
        \multirow{8}*{PointTrans.} & FGSM & 100\% & 0.8332 & 0.1544 & 0.2379 & 0.4026 \\
    ~ & 3D-ADV & 100\% & 0.3218 & \textbf{0.0006} & 0.0405 & 0.2012\\
    ~ & GeoA & 100\% & 0.4837 & 0.0185 & 0.0383 & 0.1164 \\
    & \red{AOF} & \red{100\%} & \red{0.4406} & \red{0.0278} & \red{0.0314} & \red{0.1592}\\ 
        ~ & \red{L3A} & \red{100\%} & \red{0.3192} & \red{0.0040} & \red{0.0275} & \red{0.1547} \\
        ~ & \red{SI-ADV} & \red{100\%} & \red{0.3157} & \red{0.0011} & \red{0.0379} & \red{0.1433}\\
    ~ & GSDA & 100\% & 0.1958 & 0.0073 & 0.0141 & 0.966 \\
    ~ & GSDA++  & 100\% & \textbf{0.1579} & 0.0052 & \textbf{0.0058} & \textbf{0.0822} \\ \hline
    \multirow{8}*{PointMLP} & FGSM & 100\% & 0.8029 & 0.1374 & 0.1948 & 0.3853\\
    ~ & 3D-ADV & 100\% & 0.3162 & \textbf{0.0004} & 0.0279 & 0.1895 \\
    ~ & GeoA & 100\% & 0.4578 & 0.0082 & 0.0235 & 0.0993 \\
    & \red{AOF} & \red{100\%} & \red{0.4327} & \red{0.0225} & \red{0.0271} & \red{0.1463}\\
    ~ & \red{L3A} & \red{100\%} & \red{0.3136} & \red{0.0029} & \red{0.0183} & \red{0.1404} \\
    ~ & \red{SI-ADV} & \red{100\%} & \red{0.3046} & \red{0.0009} & \red{0.0195} & \red{0.1072} \\
    ~ & GSDA & 100\% & 0.1782 & 0.0069 & 0.0090 & 0.0781 \\
    ~ & GSDA++  & 100\% & \textbf{0.1463} & 0.0006 & \textbf{0.0021} & \textbf{0.0675} \\\hline
\end{tabular}}
\label{tab:perturbation}
\end{table}

\begin{figure*}[t!]
    \centering
    \includegraphics[width=0.9\textwidth]{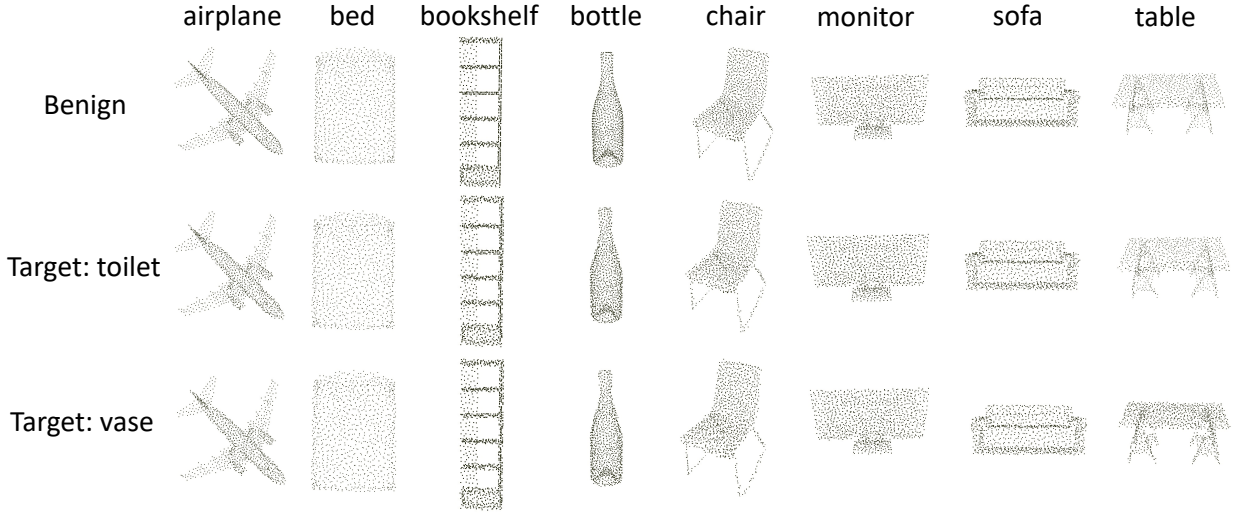}
    \caption{\new{Demonstration of representative adversarial point clouds generated by our proposed GSDA++ attack. Here, the first row shows some clean point clouds of different object classes, and the second and third rows show some corresponding target-attack results.}}
    \label{fig:all_sample}
\end{figure*}

\subsection{Evaluation on Our GSDA++ Attack}
\noindent \textbf{Quantitative Results.}
We fairly compare our GSDA++ attack with four competitive methods, including FGSM \cite{zhang2019adversarial}, 3D-ADV \cite{xiang2019generating}, GeoA \cite{wen2020geometry}, \red{AOF \cite{liu2022boosting}, L3A \cite{sun2021local}, SI-ADV \cite{huang2022shape},} and GSDA \cite{Hu2022Exploring}, and measure the perturbation in the data domain with four evaluation metrics when these methods reach 100\% of attack success rate. 
Specifically, we implement these attacks on five 3D models: PointNet, PointNet++, DGCNN, PointTrans., and PointMLP. 
The corresponding results are presented in Table~\ref{tab:perturbation}.
We see that, our GSDA++ generates adversarial point clouds with almost the lowest perturbation sizes in all evaluation metrics on five attack models.
This gives credits to our geometry-aware attack method with the proposed low-frequency constraint, achieving better imperceptibility than the other methods.
\red{AOF \cite{liu2022boosting} focuses on perturbing low-frequency components of point clouds. As low-frequency components mainly represent the basic object shape, it will result in large distortion on the original object contour. L3A \cite{sun2021local} only perturbs a subtle set of the whole point cloud, achieving better perturbation than AOF. However, this local perturbation easily introduces outliers, and thus is less imperceptible than ours. 
Different from AOF and L3A, SI-ADV \cite{huang2022shape} proposes to perturb points along the surface for improving the imperceptibility. However, it still globally perturbs the whole point cloud, including both low- and high-frequency components. Instead, we limit the perturbations within more imperceptible high-frequency components, thus achieving trivial distortion on the basic object shape represented by low-frequency components.}

\new{As for the comparison on Chamfer distance, the reason why 3D-ADV has better Chamfer distance is that the version of 3D-ADV that we compare generates a small set of independent points and places them close to the original object with only the optimization of the Chamfer distance loss. 
That is, existing points in the original point cloud are not perturbed, while a few newly synthesized points are added to the point cloud, thus leading to a lower Chamfer distance.
In comparison, our method globally attacks all points in the spectral domain for jointly preserving the geometric characteristics.
Although 3D-ADV requires the lowest perturbation sizes when measured by the Chamfer distance $\mathcal{D}_c$, it introduces noticeable outliers that result in larger local point distance $\mathcal{D}_h$, and induces much larger distortions than ours on the distance $\mathcal{D}_g$ measuring the global geometric contexts.
Besides, the reason why SI-ADV has better Chamfer distance is that, SI-ADV utilizes a complicated sensitive map to globally restrict the distances of the overall points, leading to a lower Chamfer distance. However, it still lacks awareness of the point-to-point relations, therefore resulting in larger distances in other metrics. In comparison, our attack considers utilizing spectral characteristics to perceive the point-to-point relations for better preserving the object shapes.
Moreover, from this table, we find that all attack methods take larger perturbation sizes to successfully attack PointNet++, DGCNN,  and PointTrans. than to attack PointNet and PointMLP, which indicates that PointNet++, DGCNN, and PointTrans. are harder to attack. Therefore, almost all attack methods including our attack result in larger Chamfer distance on the former three victim models while achieving lower Chamfer distance on the latter two victim models. 
As for 3D-ADV and SI-ADV, since 3D-ADV only adds a few adversarial points and SI-ADV only focuses on global distance constraints during optimization, these two methods achieve similar yet lower Chamfer distances on all five victim models. However, their performances are worse than ours in terms of other distance metrics.}
Overall, this demonstrates that our generated point clouds are less distorted quantitatively. 

\begin{figure*}[t!]
\begin{center}
    \includegraphics[width=\textwidth]{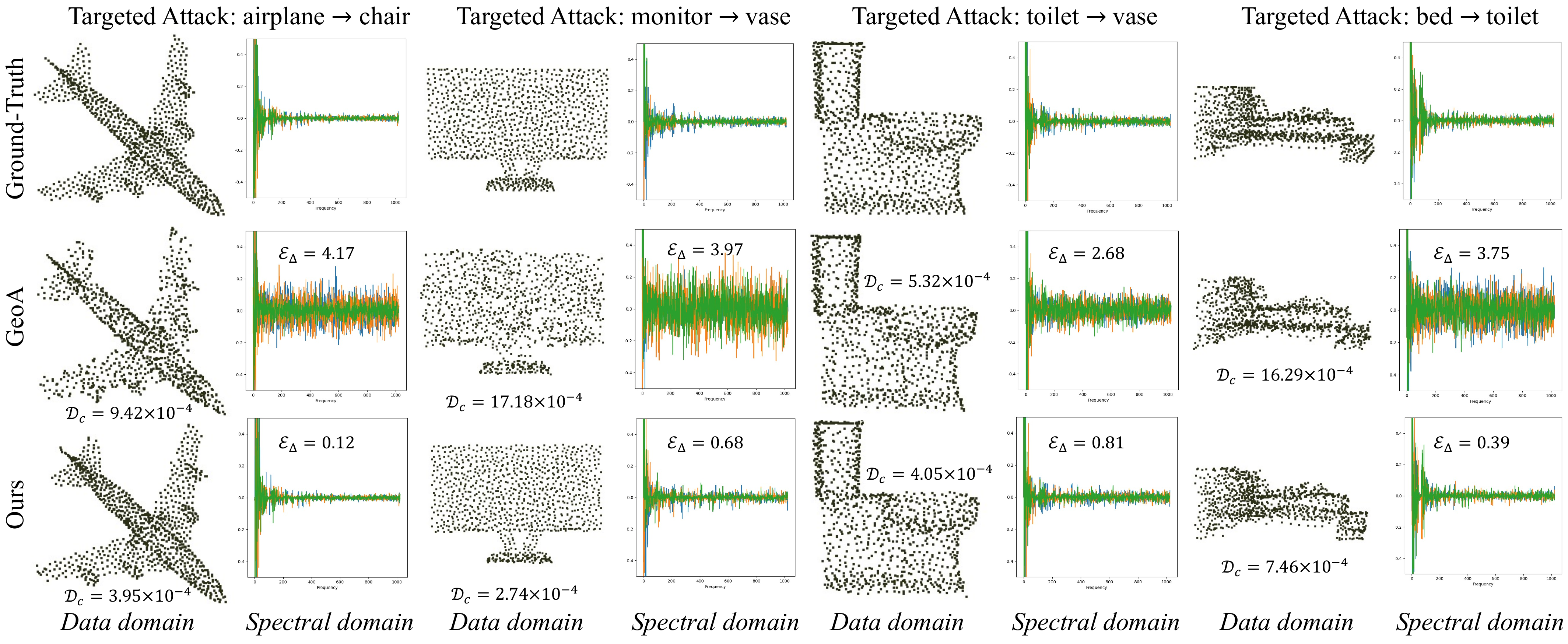}
\end{center}
\caption{Visualization of adversarial examples generated in both the data and spectral domains. Specifically, we compare our GSDA++ with GeoA using evaluation metrics of the perturbation budget $\mathcal{D}_c$ in the data domain and the perturbed energy $\mathcal{E}_{\Delta}$ in the spectral domain.}
\label{fig:visualize_contrast}
\end{figure*}

\noindent \textbf{Visualization Results.}
We show the visualization results of the generated adversarial examples in Figure~\ref{fig:all_sample}. Here, 
\new{we provide several benign point clouds in the first row, and provide corresponding representative target-attack examples in the second and third rows.}
We observe that all adversarial point clouds exhibit similar geometric structures to their corresponding benign point clouds, \ie, the attacks are quite imperceptible to humans. 
Besides, our adversarial examples have no outliers or uneven point distributions in the local area.

Also, we provide some visualization results of generated adversarial point clouds and their corresponding spectral coefficients for comparison. As shown in Figure \ref{fig:visualize_contrast}, we compare the visualization of the GeoA with ours in four examples in the context of targeted attacks. 
Since the GeoA implements perturbation in the data domain while we perform in the spectral domain, we provide the perturbation budgets in both domains for fair comparison. 
The results show that our GSDA++ attack has much less perturbation $\mathcal{E}_{\Delta}$ than the GeoA in the spectral domain, as we develop spectral noise for optimization. 
Also, by reflecting the perturbation in the data domain, our adversarial examples are more imperceptible than those of the GeoA in both local details and distributions. 
Quantitatively, we achieve a smaller perturbation budget $\mathcal{D}_c$ in the data domain. 
In conclusion, our proposed attack is more effective and imperceptible.

\subsection{Analysis on Robustness of Our GSDA++ Attack}
\label{subsec:robustness}

\begin{table}[t!]
\centering
\caption{Defense by dropping different ratios of points via SOR on the PointNet model.}
\setlength{\tabcolsep}{2.0mm}{
\begin{tabular}{cccccccc}
\hline
\multirow{3}{*}{Method} & \multicolumn{7}{c}{Attack success rate (\%)}                               \\
                        & \multicolumn{7}{c}{defense by dropping different ratios of points via SOR} \\ \cline{2-8} 
                        & 0\%     & 1\%       & 2\%      & 5\%      & 10\%     & 15\%     & 20\%     \\ \hline
GeoA                   & 100    & 83.47     & 70.56    & 52.61    & 31.58    & 18.62    & 11.71    \\
GSDA & 100 & 91.87 & 89.91 & 85.38 & 72.53 & 50.00 & 27.78 \\
GSDA++  & 100 & \textbf{93.54}     & \textbf{90.33}    & \textbf{86.75}    & \textbf{74.28}    & \textbf{53.69}    & \textbf{29.31}            \\ \hline
\end{tabular}}
\label{tab:defense_drop}
\end{table}

\begin{table}[t!]
\centering
\caption{Defense by adding different ratios of Gaussian noise on the PointNet model.}
\setlength{\tabcolsep}{2.0mm}{
\begin{tabular}{cccccccc}
\hline
\multirow{3}{*}{Method} & \multicolumn{7}{c}{Attack success rate (\%)}                               \\
                        & \multicolumn{7}{c}{adding different ratios of gaussian noise} \\ \cline{2-8} 
                        & 0\%     & 0.5\%       & 1\%      & 1.5\%      & 2\% & 2.5\% & 3\%      \\ \hline
GeoA & 100 &  86.33 & 79.26 & 76.20 & 74.12 & 70.02  & 65.17    \\
GSDA & 100 & 94.37 & 83.65 & 81.03 & 78.92 & 75.31 & 72.16  \\
GSDA++  & 100 & \textbf{100} & \textbf{99.46} & \textbf{97.35} & \textbf{95.47} & \textbf{91.11} & \textbf{85.89}               \\ \hline
\end{tabular}}
\label{tab:defense_noise}
\end{table}



\noindent \textbf{Attacking the Defenses.}
To further examine the robustness of our proposed GSDA++  attack, we employ several 3D defenses to investigate whether our attack is still effective.
Specifically, we employ the PointNet model with the following defense methods: Statistical Outlier Removal (SOR) \cite{zhou2019dup}, Simple Random Sampling (SRS) \cite{zhang2019adversarial}, \red{PointGuard \cite{liu2021pointguard},} DUP-Net defense \cite{zhou2019dup}, IF-Defense \cite{wu2020if}, \red{LPC \cite{li2022robust}, and PointDP \cite{sun2022pointdp}}. 
Table~\ref{tab:defense_drop} shows that over a range of drop ratios through the defense SOR, the attack success rates gradually drop a lot.
However, the performance of our GSDA++ attack decays much slower than that of GeoA and GSDA, thus validating that our attack is much more robust.
We also report the results with other defenses in Table~\ref{tab:various_defense}.
We observe that the FGSM and 3D-ADV attackers have low success rates under all the defenses, which is because they often lead to uneven local distribution and outliers. 
The GeoA achieves relatively higher attack success rates because it utilizes a geometry-aware loss function to constrain the similarity in the curvature and thus has fewer outliers. 
\red{
Besides, SRS, DUP-Net and IF-Defense are generally sensitive to outliers and uneven distribution. As AOF perturbs low-frequency components, L3A perturbs local points and SI-ADV perturbs points along the surface, they will result in outliers and uneven distribution and are easily recognized by these defense methods.
Instead, our GSDA++ alleviates the problems of both outliers and uneven distribution, and is thus more robust to these defenses.
LPC and PointDP defenses attempt to restore the disturbed point clouds, thus they achieve lower defense performance when it meets geometry-consistent attackers like SI-ADV and ours. However, instead of globally perturbing all points like SI-ADV, we constrain the perturbations within imperceptible high-frequency components. Hence, our GSDA++ introduces less noise than SI-ADV in the data domain, and is more robust to these defense methods than SI-ADV.
Overall}, our attack achieves the highest success rates than all other attackers under all defenses, since the trivial perturbation in the spectral domain reflects less noise in the data domain, thus enhancing the robustness.
In addition, we also investigate whether adding random Gaussian noise with a standard deviation from 0\% to 4\% of the radius of the bounding sphere would break our attack. As presented in Table~\ref{tab:defense_noise}, our adversarial examples are quite robust to such a simple defense strategy, thus demonstrating the strength of the proposed GSDA++ attack.

\begin{table}[t!]
\centering
\caption{The attack success rate (\%) on the PointNet model by various attacks under defense.}
\scalebox{0.85}{
\setlength{\tabcolsep}{0.6mm}{
\begin{tabular}{cccccccc}
\hline
Attack & No Defense & SRS & \red{PointGuard}   & DUP-Net & IF-Defense & \red{LPC} & \red{PointDP}\\ \hline
FGSM   & 100\%        & 9.68\% & \red{8.16\%}  & 4.38\%    & 4.80\%  & \red{2.45\%}  & \red{5.37\%}     \\
3D-ADV & 100\%        & 22.53\% & \red{19.21\%} & 15.44\%   & 13.70\% & \red{19.86\%}  & \red{10.79\%}   \\
\red{AOF} & \red{100\%} & \red{64.29\%} & \red{45.53\%} & \red{37.92\%} & \red{25.39\%} & \red{12.78\%} & \red{24.61\%} \\ 
\red{L3A} & \red{100\%} & \red{59.74\%} & \red{50.49\%} & \red{43.23\%} & \red{17.84\%}  & \red{33.31\%} & \red{13.53\%} \\
GeoA   & 100\%        & 67.61\% & \red{63.97\%} & 59.15\%   & 38.72\%  & \red{45.23\%}  & \red{36.02\%}  \\
\red{SI-ADV} & \red{100\%} & \red{82.47\%} & \red{72.32\%} & \red{\textbf{74.26\%}} & \red{51.55\%} & \red{56.47\%} & \red{46.45\%}  \\
GSDA   & 100\%        & 81.03\% & \red{73.85\%} & 68.98\%   & 50.26\%  & \red{60.29\%}   & \red{43.26\%}  \\
GSDA++    & 100\%        &   \textbf{83.88\%} & \red{\textbf{75.08\%}}    &  70.17\%       &   \textbf{53.64\%} & \red{\textbf{60.64\%}}  & \red{\textbf{49.94\%}}  \\ \hline
\end{tabular}}}
\label{tab:various_defense}
\end{table}

\begin{table*}[t!]
\centering
\caption{\red{The attack success rate (\%) of transfer-based attacks.}}
\setlength{\tabcolsep}{2.0mm}{
\begin{tabular}{ccccccc}
\hline
                             & \red{Attacks} & \red{PointNet} & \red{PointNet++} & \red{DGCNN} & \red{PointTrans.} & \red{PointMLP} \\ \hline
                             & \red{FGSM}    & \red{\textbf{100\%}}      & \red{3.99\%}                              & \red{0.63\%} &\red{0.57\%} &\red{1.68\%}                        \\
                             & \red{3D-ADV}  & \red{\textbf{100\%}}      & \red{8.45\%}                              & \red{1.28\%}  & \red{1.13\%} & \red{3.96\%}                       \\
                             & \red{GeoA}    & \red{\textbf{100\%}}      & \red{11.59\%}                            & \red{2.59\%}   &\red{1.97\%} &  \red{5.74\%}                      \\
\multirow{-4}{*}{\red{PointNet}}   & \red{Ours}    & \red{\textbf{100\%}} & \red{\textbf{15.27\%}} & \red{\textbf{10.30\%}}  & \red{\textbf{8.19\%}} & \red{\textbf{12.68\%}}                   \\ \hline
                             & \red{FGSM}    & \red{3.16\%}     & \red{\textbf{100\%}}                               & \red{5.57\%}  & \red{4.63\%} & \red{2.89\%}                       \\
                             & \red{3D-ADV}  & \red{6.63\%}     & \red{\textbf{100\%}}                               & \red{10.98\%}  & \red{10.12\%} & \red{5.74\%}                      \\
                             & \red{GeoA}    & \red{9.47\%}     & \red{\textbf{100\%}}                               & \red{19.77\%} & \red{16.81\%} & \red{7.38\%}                       \\
\multirow{-4}{*}{\red{PointNet++}} & \red{Ours} & \red{\textbf{13.71\%}} & \red{\textbf{100\%}} & \red{\textbf{31.62\%}} & \red{\textbf{26.54\%}} & \red{\textbf{18.35\%}}          \\ \hline
                             & \red{FGSM}    & \red{3.59\%}     & \red{7.21\%}                              & \red{\textbf{100\%}} & \red{5.79\%} & \red{3.17\%}                         \\
                             & \red{3D-ADV}  & \red{6.82\%}     & \red{13.53\%}                             & \red{\textbf{100\%}} & \red{11.84\%} & \red{6.26\%}                          \\
                             & \red{GeoA}    & \red{12.46\%}    & \red{24.24\%}                             & \red{\textbf{100\%}}   & \red{19.40\%} & \red{10.15\%}                       \\
\multirow{-4}{*}{\red{DGCNN}}      & \red{Ours} & \red{\textbf{33.74\%}} & \red{\textbf{84.79\%}} & \red{\textbf{100\%}} & \red{\textbf{71.48\%}} & \red{\textbf{31.03\%}}                 \\ \hline
\multirow{4}*{\red{PointTrans.}} & \red{FGSM} & \red{3.25\%} & \red{7.38\%} & \red{6.43\%} & \red{\textbf{100\%}} & \red{3.51\%} \\
    ~ & \red{3D-ADV} & \red{6.54\%} & \red{13.60\%} & \red{13.02\%} & \red{\textbf{100\%}} & \red{6.89\%} \\
    ~ & \red{GeoA} & \red{12.57\%} & \red{23.64\%} & \red{22.15\%} & \red{\textbf{100\%}} & \red{12.29\%} \\
    ~ & \red{Ours}  & \red{\textbf{32.15\%}} & \red{\textbf{79.62\%}} & \red{\textbf{75.90\%}} & \red{\textbf{100\%}} & \red{\textbf{29.83\%}}\\ \hline
    {\multirow{4}*{\red{PointMLP}}} & \red{FGSM} & \red{4.35\%} & \red{3.81\%} & \red{1.74\%} & \red{1.09\%} & \red{\textbf{100\%}} \\
    ~ & \red{3D-ADV} & \red{9.27\%} & \red{8.60\%} & \red{4.52\%} & \red{3.54\%} & \red{\textbf{100\%}} \\
    ~ & \red{GeoA} & \red{12.38\%} & \red{11.63\%} & \red{6.75\%} & \red{5.72\%} & \red{\textbf{100\%}} \\
    ~ & \red{Ours} & \red{\textbf{27.28\%}} & \red{\textbf{23.49\%}} & \red{\textbf{15.51\%}} & \red{\textbf{12.32\%}} & \red{\textbf{100\%}}\\ \hline
\end{tabular}}
\label{tab:transfer}
\end{table*}

\noindent \red{\textbf{Transferability of Adversarial Point Clouds.}}
\red{
To investigate the transferability of our proposed GSDA++ attack, we craft adversarial point clouds on normally trained models and directly test them on all the five 3D models we consider. The success rates, which are the misclassification rates of the corresponding models on adversarial examples, are shown in Table~\ref{tab:transfer}. The rows and columns of the table present the models we attack and the five models we test.
We observe that our GSDA++ attack has relatively higher success rates of transfer-based attacks than others. 
This is because our perturbation in the spectral domain is able to keep the spectral characteristics of 3D objects, leading to imperceptible adversarial samples with trivial noise in the data domain. Therefore, our attack is more robust to unknown defenses or transformations in unseen victim models.}

\noindent \textbf{Properties of Our GSDA++ Attack.}
We conduct experiments to validate the properties of robustness to scaling and rotation. 
As shown in Table~\ref{tab:our_invariance}, applying scaling (randomly scale the point cloud by a ratio in the range $[0.8,1.2]$) and rotation (randomly rotate the point cloud by an angle in the range $[-180^{\circ},180^{\circ}]$) to existing attacks (FGSM, 3D-ADV, and GeoA) greatly degenerates the attack performance. 
In contrast, our attack is more robust to these transformations and still achieves a high attack success rate.

\begin{table}[t!]
\centering
\caption{Investigate the properties of our GSDA++ attack on PointNet. The values are the attack success rate (\%).}
\setlength{\tabcolsep}{1.6mm}{
\begin{tabular}{ccccc}
\hline
Operation & FGSM & 3D-ADV & GeoA & GSDA++ \\ \hline
None & \textbf{100\%} & \textbf{100\%} & \textbf{100\%} & \textbf{100\%} \\
Scaling & 29.65\% & 34.82\% & 39.73\% & \textbf{84.16\%}\\
Rotation & 43.74\% & 48.31\% & 51.08\% & \textbf{87.69\%} 
\\ \hline
\end{tabular}}
\label{tab:our_invariance}
\end{table}

\begin{table}[t!]
\centering
\caption{The attack success rate (\%) on different models by various attacks under our proposed defense strategy. Here, we only use clean samples that are 100\% accurately classified by the 3D models.}
\setlength{\tabcolsep}{1.2mm}{
\begin{tabular}{ccccc}
\hline
Attack & Defense & PointNet & PointNet++  & DGCNN \\ \hline
\multirow{2}*{GeoA} & w/o defense & 100\% & 100\% & 100\%\\
~ & w/ defense & 64.85\% & 69.33\% & 70.61\%
\\ \hline
\multirow{2}*{GSDA} & w/o defense & 100\% & 100\% & 100\%\\
~ & w/ defense & 38.26\% & 40.12\% & 34.49\%
\\ \hline
\multirow{2}*{GSDA++} & w/o defense & 100\% & 100\% & 100\%\\
~ & w/ defense & \textbf{12.58\%} & \textbf{8.87\%} & \textbf{11.25\%} 
\\ \hline
\end{tabular}}
\label{tab:our_defense}
\end{table}

\begin{table}[t!]
    \centering
    \caption{\red{The accuracy of the model on clean data. * denotes the model trained with our defense method.}}
    \renewcommand{\arraystretch}{1.2}{
    \setlength{\tabcolsep}{0.6mm}{
    \begin{tabular}{c|ccccc}
    \hline
     & \red{PointNet} &  \red{PointNet++} &  \red{DGCNN} &  \red{PointTrans.} &  \red{PointMLP} \\ \hline
      \red{Accuracy} & \textbf{ \red{89.2\%}} & \red{91.9\%} &  \red{92.2\%} &  \red{93.7\%} & \textbf{\red{94.5\%}} \\ \hline
     & \red{PointNet*} & \red{PointNet++*} & \red{DGCNN*} & \red{PointTrans.*} & \red{PointMLP*} \\ \hline
     \red{Accuracy} & \red{89.1\%} & \textbf{\red{92.2\%}} & \textbf{\red{92.4\%}} & \textbf{\red{94.0\%}} & \red{94.1\%} \\ \hline
    \end{tabular}}}
    \label{tab:clean}
\end{table}


\subsection{Analysis on Robustness of Our Defense Strategy}
To verify the effectiveness of our proposed low-frequency-aware defense method, we first train victim models from scratch with the training samples of both clean data and their low-pass versions, and then measure the attack success rates. 
As shown in Table~\ref{tab:our_defense}, we evaluate the effectiveness of our defense method on three commonly adopted 3D models, respectively. 
From this table, we have the following observation:
1) The results show that the proposed training strategy defends our proposed GSDA++ attack well, demonstrating that recognizing the original object shape represented by its low-frequency components is robust to the adversarial noise in mid-high frequency components.
2) Our defense strategy is also robust to other spectral attacks like GSDA, since its spectral perturbations are learned to generate less spectral noise on the low-frequency band for preserving the geometric characteristics.
3) Our defense performs relatively worse on data-domain attacks like GeoA, since such attacker fails to capture the spectral contexts.

\red{Besides, to investigate whether our spectrum-aware adversarial training technique affects the classification accuracy on clean data, we report the model performance on the whole test set of clean point clouds. 
The accuracy of the model trained with and without our defense method is competitive, as shown in Table~\ref{tab:clean}. 
This verifies that our training achieves stronger robustness while maintaining the accuracy over clean data.}

\subsection{Ablation Study}

\begin{table}[t!]
\centering
\caption{Comparison of using different spectral representations on PointNet.}
\setlength{\tabcolsep}{1.4mm}{
\begin{tabular}{cccccccc}
\hline
\multirow{3}{*}{Method} & \multicolumn{4}{c}{Attack success rate (\%)} & \multirow{3}{*}{$\mathcal{D}_{c}$} & \multirow{3}{*}{$\mathcal{D}_{h}$} & \multirow{3}{*}{$\mathcal{E}_{\Delta}$}   \\
                        & \multicolumn{4}{c}{defense via SOR}          &                     &                     &                      \\ \cline{2-5}
                        & 0\%      & 5\%       & 10\%      & 20\%      &                     &                     &                      \\ \hline
Ours-DCT & 92.24 & 3.79 & 0.46 & 0.12 & 0.0038 & 0.0397 & 32.4785 \\
Ours-GFT                    & \textbf{100}      & \textbf{86.75}     & \textbf{74.28}     & \textbf{29.31}     & \textbf{0.0006}              & \textbf{0.0028}              & \textbf{0.7580}  \\ \hline
\end{tabular}}
\label{tab:dct}
\end{table}

\subsubsection{Effect of the GFT Transform}
We first evaluate the benefit of spectral representation in the GFT domain.  
Since the DCT is widely adopted in the {\it regular} field, such as transforming images onto the spectral domain, we compare the spectral domain attacks on different spectral representations---the DCT and the GFT. 
In order to represent an irregular point cloud via the DCT, we quantize the point cloud into regular 3D voxels and implement the attack in the DCT domain via the regular 3D-DCT, which is denoted as Ours-DCT.
As listed in Table~\ref{tab:dct}, our attack with the GFT is more robust to the SOR defense and requires much lower perturbation size in both the data domain and spectral domain than the attack with the DCT.
The main reason is that point clouds are unordered, which makes it challenging to capture the correlations among points by the DCT.
Besides, the voxelization operation inevitably introduces quantization error.
In contrast, the GFT captures the underlying structure of point clouds well via the appropriate graph construction.
This validates our choice of the GFT for compact representations of point clouds in the spectral domain. 

\begin{table}[t!]
\centering
\caption{Sensitivity analysis of the number $K$.}
\begin{tabular}{ccccc}
\hline
Number $K$ & Success Rate & $\mathcal{D}_c$ & $\mathcal{D}_h$ & $\mathcal{E}_{\Delta}$ \\ \hline
$K$=5 & 100\% & \textbf{0.0006}  & 0.0032 & 0.9247 \\
$K$=10 & 100\% & \textbf{0.0006} & 0.0028 & \textbf{0.7580} \\
$K$=20 & 100\% & \textbf{0.0006} & \textbf{0.0027} & 0.7643 \\
$K$=40 & 100\% & \textbf{0.0006} & \textbf{0.0027} & 0.7685 \\ \hline
\end{tabular}
\label{tab:ablation_K}
\end{table}

\subsubsection{Sensitivity on the Number $K$}
As shown in Table~\ref{tab:ablation_K}, we investigate whether the adversarial effects vary with respect to different settings of the number $K$ in the $K$-NN graph.
Specifically, we test on $K=5,10,20,40$ respectively to perform the proposed GSDA++ attack on the PointNet model and report the corresponding perturbation budgets when achieving 100\% of the attack success rate. 
We see that, the attack performance is insensitive to $K$ since our attack with different $K$'s requires similar perturbation budgets $\mathcal{D}_c,\mathcal{D}_h$ in the data domain, as well as in the spectral domain measured by $\mathcal{E}_{\Delta}$.
Therefore, we set $K=10$ in all our experiments.

\begin{figure}[t!]
\begin{center}
    \includegraphics[width=0.48\textwidth]{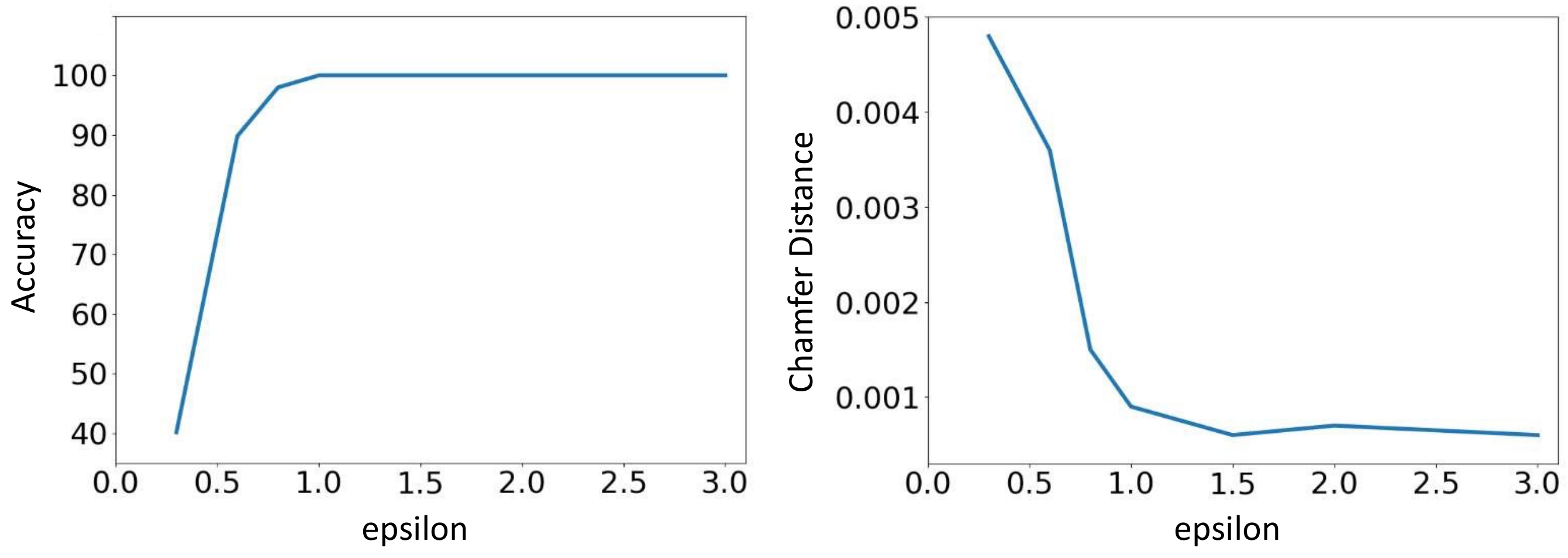}
\end{center}
\caption{\red{Attacking performance when applying different restrictions by $\epsilon$, in terms of success rate, Chamfer distance between adversarial point clouds and the originals, on the PointNet model.}}
\label{fig:epsilon}
\end{figure}

\subsubsection{\red{Sensitivity on the perturbation size $\epsilon$}}
\red{The parameter $\epsilon$ restricts the perturbation size in the spectral domain, which aims to preserve the original spectral characteristics so that the resultant adversarial point cloud is visually indistinguishable from its clean version.
As shown in Figure~\ref{fig:epsilon}, we conduct the ablation study on
the perturbation constraint $\epsilon$ in the spectral domain with different values on the ModelNet40 dataset. It shows that the success rates of our GSDA++ improve with the increase of $\epsilon$, and achieve 100\% of success rate when $\epsilon=1.5$. Meanwhile, the Chamfer distance is small enough when $\epsilon=1.5$. Therefore, we set $\epsilon=1.5$ in all the experiments.}

\begin{table}[t!]
\centering
\caption{\red{Ablation study on parameters of frequency band on PointNet model.}}
\begin{tabular}{c|ccc}
\hline
\red{Energy Split} & \red{$\mathcal{D}_c$} & \red{$\mathcal{D}_h$} & \red{$\mathcal{E}_{\Delta}$} \\ \hline
\red{(75\%,90\%)} & \red{0.0006} & \red{0.0028} & \red{\textbf{0.7580}} \\ \hline
\red{(73\%,90\%)} & \red{0.0006} &\red{0.0030} & \red{0.7617} \\
\red{(77\%,90\%)} & \red{0.0006} & \red{\textbf{0.0027}} & \red{0.7602}\\ 
\red{(75\%,88\%)} & \red{\textbf{0.0005}} & \red{0.0029} & \red{0.7584} \\ 
\red{(75\%,92\%)} & \red{0.0009} & \red{0.0038} & \red{0.7633}\\
\hline
\end{tabular}
\label{tab:para}
\end{table}

\subsubsection{\red{Sensitivity on the frequency band}}
\red{As investigated in Sec. 3.2.2, we evaluate the spectral characteristics on 12311 point clouds, and find that they have almost 75\% of energy within the lowest 100 frequencies and almost 90\% of energy within the lowest 400 frequencies. 
Based on this observation, we set three frequency bands for point clouds with 1024 points on the ModelNet40 dataset: the low-frequency band (frequency $[0,100)$), the mid-frequency band (frequency $[100,400)$), and the high-frequency band (frequency $[400,1024]$).
We have implemented the case study of slightly changing on this parameter, as shown in Table~\ref{tab:para}. We find that the performance is insensitive to slight changes over this parameter.}

\begin{table}[t!]
\centering
\caption{Ablation study on the \new{graph spectral filter}.}
\begin{tabular}{c|ccc|ccc}
\hline
Variant & $\bm{\Delta}_{w}$ & $\bm{\Delta}_{h}$ & L & $\mathcal{D}_c$ & $\mathcal{D}_h$ & $\mathcal{E}_{\Delta}$ \\ \hline
1. & $\checkmark$ & $\times$ & - & 0.0009 & 0.0047 & 1.3428 \\
2. & $\times$ & $\checkmark$ & 5 & 0.0149 & 0.0203 & 3.0162 \\
3. & $\checkmark$ & $\checkmark$ & 5 & \textbf{0.0006} & \textbf{0.0028} & 0.7580 \\ \hline
4. & $\checkmark$ & $\checkmark$ & 1 & 0.0008 & 0.0043 & 1.1526 \\
5. & $\checkmark$ & $\checkmark$ & 3 & 0.0007 & 0.0032 & 0.8075 \\
6. & $\checkmark$ & $\checkmark$ & 7 & \textbf{0.0006} & 0.0031 & \textbf{0.7534} \\ \hline
\end{tabular}
\label{tab:variant}
\end{table}

\subsubsection{Design of the Graph Spectral Filter}
The proposed graph spectral filter in Eq.~(\ref{eq:objective}) consists of two kinds of learnable parameters: 
1) the coefficients of the polynomial function at each order, $\bm{\Delta}_{h,l},l=0,...,L-1$; 
2) the multiplier before each polynomial function, $\bm{\Delta}_{w,i},i=1,...,n$, 
\red{Here, we utilize both of them to learn a desirable spectral distribution for better attack.
Specifically, $\bm{\Delta}_{h,l},l=0,...,L-1$ denotes the coefficients of the polynomial function at each order for learning a desirable distribution for attack, and $\bm{\Delta}_{w,i},i=1,...,n$
learns the multiplier before each polynomial function for highlighting the contributed frequency components. We conduct ablation studies on them and present the results in  Table~\ref{tab:variant}.
}
We have the following observations:
1) \red{Comparing variants 1-3, we find that adding $\Delta_h$ to variant 1 brings the improvement of 0.0003 on $\mathcal{D}_c$, while adding $\Delta_w$ to variant 2 brings the improvement of 0.0143 on $\mathcal{D}_c$.}
2) Comparing variants 3-6, the order of the polynomial function should be appropriately chosen so as to fit the desirable distribution well. Experimentally, the model achieves the best performance when $L=5$.
\red{3) Both of $\Delta_w$ and $\Delta_h$ contribute substantially to fit the desirable distribution for imperceptibly attacking victim models in the spectral domain.}

\begin{table}[t!]
\centering
\caption{Comparative results on the perturbation sizes of GSDA++ without or with the low-frequency constraint (LFC).}
\setlength{\tabcolsep}{1.6mm}{
\begin{tabular}{cccccc}
\hline
\multirow{2}{*}{\begin{tabular}[c]{@{}c@{}}Attack\\ Model\end{tabular}} & \multirow{2}{*}{Variant} & \multicolumn{4}{c}{Perturbation Size} \\ \cline{3-6} 
&                          &                                                                      $\mathcal{D}_{norm}$     & $\mathcal{D}_{c}$       & $\mathcal{D}_{h}$ & $\mathcal{D}_{g}$      \\ \hline
\multirow{2}{*}{PointNet} & w/o. LFC & 0.1975 & 0.0029 & 0.0071 & 0.0793 \\                                             ~ & w/. LFC                                                                                       &     \textbf{0.1517}        & \textbf{0.0006}              & \textbf{0.0028} & \textbf{0.0633}    \\ \hline
\multirow{2}{*}{PointNet++}      
& w/o. LFC & 0.2031 & 0.0102 & 0.0170 & 0.1066 \\                                   ~    & w/. LFC                                                                                         &     \textbf{0.1664}        &     0.0065       & \textbf{0.0128} & \textbf{0.0986}           \\  \hline
\multirow{2}{*}{DGCNN}  & w/o. LFC & 0.2062 & 0.0164 & 0.0218 & 0.1083 \\                                        ~        & w/. LFC                                                                                       &     \textbf{0.1731}        & 0.0072     & \textbf{0.0135} & \textbf{0.0960}    \\  \hline
        \multirow{2}*{PointTrans.} & w/o. LFC & 0.1885  & 0.0119 & 00103 & 0.0930 \\
        ~ & w/. LFC   & \textbf{0.1579} & 0.0052 & \textbf{0.0058} & \textbf{0.0822} \\ 
       \hline
    \multirow{2}*{PointMLP} 
    & w/o. LFC & 0.1826 & 0.0025 & 0.0069 & 0.0824\\ 
    ~ & w/. LFC  & \textbf{0.1463} & 0.0006 & \textbf{0.0021} & \textbf{0.0675} \\\hline
\end{tabular}}
\label{tab:low_frequency}
\end{table}

\begin{figure}[t!]
\begin{center}
    \includegraphics[width=0.48\textwidth]{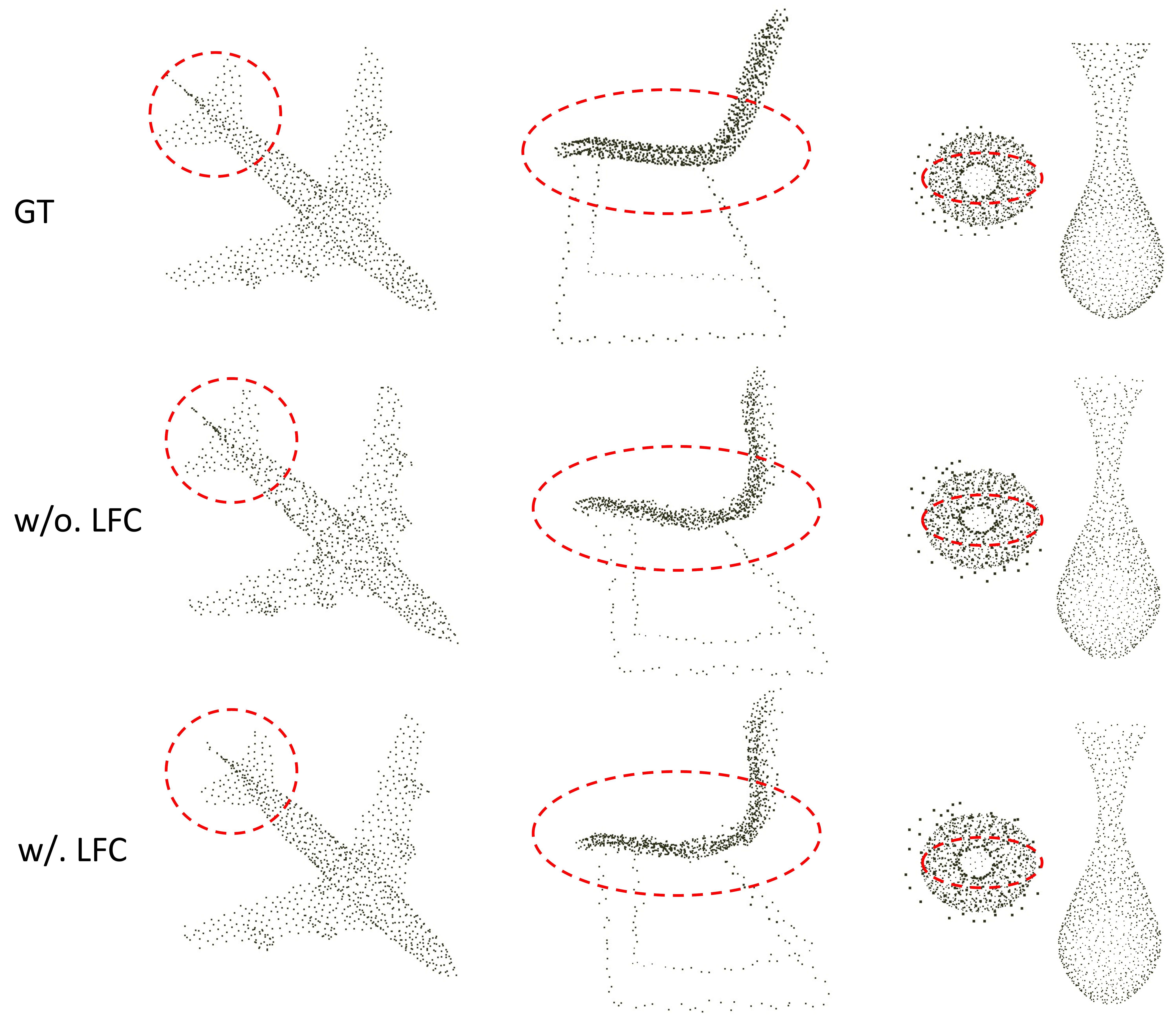}
\end{center}
\caption{Visualization results on adversarial examples without or with the low-frequency constraint (LFC).}
\label{fig:low_frequency}
\end{figure}

\subsubsection{Effect of the Low-Frequency Constraint}
To investigate the effectiveness of the proposed low-frequency constraint (LFC) during the spectral attack, we evaluate and compare the adversarial point clouds generated with and without the LFC.
As shown in Figure~\ref{fig:low_frequency}, without the LFC design, the adversarial perturbations are reasonable and are imperceptible to humans, as there is no uneven point distribution or outlier in general. 
This shows that attacking point clouds in the spectral domain preserves the characteristics of piecewise smoothness, thus leading to high-quality imperceptible adversarial samples.
Moreover, by applying the additional LFC, the adversarial examples exhibit smaller geometric distortion compared to the original ones, further improving the imperceptibility, such as around the legs of the chair.
We also measure the distance metrics of both variants in Table~\ref{tab:low_frequency}. 
The results show that the adversarial point clouds generated with the LFC require smaller perturbation sizes, since the prominent structure is preserved via the constraint over the low-frequency components.

\section{Conclusion}
In this paper, we propose a novel paradigm of point cloud attacks---Graph Spectral Domain Attack (GSDA++), which explores insightful spectral characteristics and performs perturbation in the spectral domain via a learnable graph spectral filter to generate adversarial examples with well-preserved geometric structures. 
In addition, considering that the low-frequency components of point clouds mainly characterize the basic shape of the 3D object, we propose a low-frequency constraint that guides the perturbation to concentrate on the high-frequency components. 
Also, we provide a simple yet effective defense strategy to defend such spectral attacks.
Extensive experiments demonstrate the vulnerability of popular 3D models to the proposed GSDA++ and validate the robustness of our adversarial point clouds.

\bibliographystyle{IEEEtran}
\bibliography{reference.bib}

\begin{thebibliography}{10}
\providecommand{\url}[1]{#1}
\csname url@samestyle\endcsname
\providecommand{\newblock}{\relax}
\providecommand{\bibinfo}[2]{#2}
\providecommand{\BIBentrySTDinterwordspacing}{\spaceskip=0pt\relax}
\providecommand{\BIBentryALTinterwordstretchfactor}{4}
\providecommand{\BIBentryALTinterwordspacing}{\spaceskip=\fontdimen2\font plus
\BIBentryALTinterwordstretchfactor\fontdimen3\font minus \fontdimen4\font\relax}
\providecommand{\BIBforeignlanguage}[2]{{%
\expandafter\ifx\csname l@#1\endcsname\relax
\typeout{** WARNING: IEEEtran.bst: No hyphenation pattern has been}%
\typeout{** loaded for the language `#1'. Using the pattern for}%
\typeout{** the default language instead.}%
\else
\language=\csname l@#1\endcsname
\fi
#2}}
\providecommand{\BIBdecl}{\relax}
\BIBdecl

\bibitem{goodfellow2014explaining}
I.~J. Goodfellow, J.~Shlens, and C.~Szegedy, ``Explaining and harnessing adversarial examples,'' \emph{arXiv preprint arXiv:1412.6572}, 2014.

\bibitem{szegedy2013intriguing}
C.~Szegedy, W.~Zaremba, I.~Sutskever, J.~Bruna, D.~Erhan, I.~Goodfellow, and R.~Fergus, ``Intriguing properties of neural networks,'' \emph{arXiv preprint arXiv:1312.6199}, 2013.

\bibitem{dong2018boosting}
Y.~Dong, F.~Liao, T.~Pang, H.~Su, J.~Zhu, X.~Hu, and J.~Li, ``Boosting adversarial attacks with momentum,'' in \emph{Proceedings of the IEEE Conference on Computer Vision and Pattern Recognition (CVPR)}, 2018, pp. 9185--9193.

\bibitem{madry2017towards}
A.~Madry, A.~Makelov, L.~Schmidt, D.~Tsipras, and A.~Vladu, ``Towards deep learning models resistant to adversarial attacks,'' \emph{arXiv preprint arXiv:1706.06083}, 2017.

\bibitem{kurakin2016adversarial}
A.~Kurakin, I.~Goodfellow, and S.~Bengio, ``Adversarial machine learning at scale,'' \emph{arXiv preprint arXiv:1611.01236}, 2016.

\bibitem{tu2019autozoom}
C.-C. Tu, P.~Ting, P.-Y. Chen, S.~Liu, H.~Zhang, J.~Yi, C.-J. Hsieh, and S.-M. Cheng, ``Autozoom: Autoencoder-based zeroth order optimization method for attacking black-box neural networks,'' in \emph{Proceedings of the AAAI Conference on Artificial Intelligence}, vol.~33, no.~01, 2019, pp. 742--749.

\bibitem{chen2017multi}
X.~Chen, H.~Ma, J.~Wan, B.~Li, and T.~Xia, ``Multi-view 3d object detection network for autonomous driving,'' in \emph{Proceedings of the IEEE conference on Computer Vision and Pattern Recognition (CVPR)}, 2017, pp. 1907--1915.

\bibitem{singh20203d}
S.~P. Singh, L.~Wang, S.~Gupta, H.~Goli, P.~Padmanabhan, and B.~Guly{\'a}s, ``3d deep learning on medical images: a review,'' \emph{Sensors}, vol.~20, no.~18, p. 5097, 2020.

\bibitem{xiang2019generating}
C.~Xiang, C.~R. Qi, and B.~Li, ``Generating 3d adversarial point clouds,'' in \emph{Proceedings of the IEEE Conference on Computer Vision and Pattern Recognition (CVPR)}, 2019, pp. 9136--9144.

\bibitem{wicker2019robustness}
M.~Wicker and M.~Kwiatkowska, ``Robustness of 3d deep learning in an adversarial setting,'' in \emph{Proceedings of the IEEE Conference on Computer Vision and Pattern Recognition (CVPR)}, 2019, pp. 11\,767--11\,775.

\bibitem{zhang2019adversarial}
Q.~Zhang, J.~Yang, R.~Fang, B.~Ni, J.~Liu, and Q.~Tian, ``Adversarial attack and defense on point sets,'' \emph{arXiv preprint arXiv:1902.10899}, 2019.

\bibitem{zheng2019pointcloud}
T.~Zheng, C.~Chen, J.~Yuan, B.~Li, and K.~Ren, ``Pointcloud saliency maps,'' in \emph{Proceedings of the IEEE International Conference on Computer Vision (ICCV)}, 2019, pp. 1598--1606.

\bibitem{tsai2020robust}
T.~Tsai, K.~Yang, T.-Y. Ho, and Y.~Jin, ``Robust adversarial objects against deep learning models,'' in \emph{Proceedings of the AAAI Conference on Artificial Intelligence}, vol.~34, no.~01, 2020, pp. 954--962.

\bibitem{zhao2020isometry}
Y.~Zhao, Y.~Wu, C.~Chen, and A.~Lim, ``On isometry robustness of deep 3d point cloud models under adversarial attacks,'' in \emph{Proceedings of the IEEE Conference on Computer Vision and Pattern Recognition (CVPR)}, 2020, pp. 1201--1210.

\bibitem{zhou2020lg}
H.~Zhou, D.~Chen, J.~Liao, K.~Chen, X.~Dong, K.~Liu, W.~Zhang, G.~Hua, and N.~Yu, ``Lg-gan: Label guided adversarial network for flexible targeted attack of point cloud based deep networks,'' in \emph{Proceedings of the IEEE Conference on Computer Vision and Pattern Recognition (CVPR)}, 2020, pp. 10\,356--10\,365.

\bibitem{hamdi2020advpc}
A.~Hamdi, S.~Rojas, A.~Thabet, and B.~Ghanem, ``Advpc: Transferable adversarial perturbations on 3d point clouds,'' in \emph{European Conference on Computer Vision (ECCV)}, 2020, pp. 241--257.

\bibitem{wen2020geometry}
Y.~Wen, J.~Lin, K.~Chen, C.~P. Chen, and K.~Jia, ``Geometry-aware generation of adversarial point clouds,'' \emph{IEEE Transactions on Pattern Analysis and Machine Intelligence (TPAMI)}, 2020.

\bibitem{carlini2017towards}
N.~Carlini and D.~Wagner, ``Towards evaluating the robustness of neural networks,'' in \emph{2017 IEEE Symposium on Security and Privacy (SP)}, 2017, pp. 39--57.

\bibitem{liu2019extending}
D.~Liu, R.~Yu, and H.~Su, ``Extending adversarial attacks and defenses to deep 3d point cloud classifiers,'' in \emph{2019 IEEE International Conference on Image Processing (ICIP)}, 2019, pp. 2279--2283.

\bibitem{ma2020efficient}
C.~Ma, W.~Meng, B.~Wu, S.~Xu, and X.~Zhang, ``Efficient joint gradient based attack against sor defense for 3d point cloud classification,'' in \emph{Proceedings of the 28th ACM International Conference on Multimedia}, 2020, pp. 1819--1827.

\bibitem{zhang2019defense}
Y.~Zhang, G.~Liang, T.~Salem, and N.~Jacobs, ``Defense-pointnet: Protecting pointnet against adversarial attacks,'' in \emph{2019 IEEE International Conference on Big Data (Big Data)}, 2019, pp. 5654--5660.

\bibitem{hu2021overview}
W.~Hu, J.~Pang, X.~Liu, D.~Tian, C.-W. Lin, and A.~Vetro, ``Graph {S}ignal {P}rocessing for geometric data and beyond: Theory and applications,'' \emph{IEEE Transactions on Multimedia}, 2021.

\bibitem{hu2014multiresolution}
W.~Hu, G.~Cheung, A.~Ortega, and O.~C. Au, ``Multiresolution graph {F}ourier transform for compression of piecewise smooth images,'' \emph{IEEE Trans. Image Process.}, vol.~24, no.~1, pp. 419--433, 2015.

\bibitem{chao2015edge}
Y.-H. Chao, A.~Ortega, W.~Hu, and G.~Cheung, ``Edge-adaptive depth map coding with lifting transform on graphs,'' in \emph{2015 Picture Coding Symposium (PCS)}.\hskip 1em plus 0.5em minus 0.4em\relax IEEE, 2015, pp. 60--64.

\bibitem{choi2020task}
J.~Choi and B.~Han, ``Task-aware quantization network for jpeg image compression,'' in \emph{European Conference on Computer Vision (ECCV)}, 2020, pp. 309--324.

\bibitem{li2018learning}
M.~Li, W.~Zuo, S.~Gu, D.~Zhao, and D.~Zhang, ``Learning convolutional networks for content-weighted image compression,'' in \emph{Proceedings of the IEEE Conference on Computer Vision and Pattern Recognition (CVPR)}, 2018, pp. 3214--3223.

\bibitem{hammond2011wavelets}
D.~K. Hammond, P.~Vandergheynst, and R.~Gribonval, ``Wavelets on graphs via spectral graph theory,'' \emph{Appl. Comput. Harmonic Anal.}, vol.~30, no.~2, pp. 129--150, 2011.

\bibitem{zhang2012analyzing}
C.~Zhang and D.~Flor{\^e}ncio, ``Analyzing the optimality of predictive transform coding using graph-based models,'' \emph{IEEE Signal Processing Letters}, vol.~20, no.~1, pp. 106--109, 2012.

\bibitem{shen2010edge}
G.~Shen, W.-S. Kim, S.~K. Narang, A.~Ortega, J.~Lee, and H.~Wey, ``Edge-adaptive transforms for efficient depth map coding,'' in \emph{Proc. Picture Coding Symp.}, 2010, pp. 566--569.

\bibitem{hu2012depth}
W.~Hu, G.~Cheung, X.~Li, and O.~Au, ``Depth map compression using multi-resolution graph-based transform for depth-image-based rendering,'' in \emph{Proc. IEEE Int. Conf. Image Process.}, 2012, pp. 1297--1300.

\bibitem{hu2015intra}
W.~Hu, G.~Cheung, and A.~Ortega, ``Intra-prediction and generalized graph {F}ourier transform for image coding,'' \emph{IEEE Signal Process. Lett.}, vol.~22, no.~11, pp. 1913--1917, 2015.

\bibitem{zhang2014point}
C.~Zhang, D.~Florencio, and C.~Loop, ``Point cloud attribute compression with graph transform,'' in \emph{Proc. IEEE Int. Conf. Image Process.}, 2014, pp. 2066--2070.

\bibitem{xu2020predictive}
Y.~Xu, W.~Hu, S.~Wang, X.~Zhang, S.~Wang, S.~Ma, Z.~Guo, and W.~Gao, ``Predictive generalized graph fourier transform for attribute compression of dynamic point clouds,'' \emph{IEEE Transactions on Circuits and Systems for Video Technology}, vol.~31, no.~5, pp. 1968--1982, 2020.

\bibitem{shuman2013emerging}
D.~I. Shuman, S.~K. Narang, P.~Frossard, A.~Ortega, and P.~Vandergheynst, ``The emerging field of signal processing on graphs: {Extending} high-dimensional data analysis to networks and other irregular domains,'' \emph{IEEE Signal Process. Mag.}, vol.~30, no.~3, pp. 83--98, 2013.

\bibitem{Hu2022Exploring}
Q.~Hu, D.~Liu, and W.~Hu, ``Exploring the devil in graph spectral domain for 3d point cloud attacks,'' in \emph{European Conference on Computer Vision (ECCV)}, 2022.

\bibitem{qi2017pointnet}
C.~R. Qi, H.~Su, K.~Mo, and L.~J. Guibas, ``Pointnet: Deep learning on point sets for 3d classification and segmentation,'' in \emph{Proceedings of the IEEE Conference on Computer Vision and Pattern Recognition (CVPR)}, 2017, pp. 652--660.

\bibitem{qi2017pointnet++}
C.~R. Qi, L.~Yi, H.~Su, and L.~J. Guibas, ``Pointnet++: Deep hierarchical feature learning on point sets in a metric space,'' \emph{Advances in Neural Information Processing Systems (NIPS)}, 2017.

\bibitem{wang2019dynamic}
Y.~Wang, Y.~Sun, Z.~Liu, S.~E. Sarma, M.~M. Bronstein, and J.~M. Solomon, ``Dynamic graph cnn for learning on point clouds,'' \emph{Acm Transactions On Graphics (TOG)}, vol.~38, no.~5, pp. 1--12, 2019.

\bibitem{yang2018foldingnet}
Y.~Yang, C.~Feng, Y.~Shen, and D.~Tian, ``Foldingnet: Point cloud auto-encoder via deep grid deformation,'' in \emph{Proceedings of the IEEE Conference on Computer Vision and Pattern Recognition (CVPR)}, 2018, pp. 206--215.

\bibitem{liu2019relation}
Y.~Liu, B.~Fan, S.~Xiang, and C.~Pan, ``Relation-shape convolutional neural network for point cloud analysis,'' in \emph{Proceedings of the IEEE Conference on Computer Vision and Pattern Recognition (CVPR)}, 2019, pp. 8895--8904.

\bibitem{rao2020global}
Y.~Rao, J.~Lu, and J.~Zhou, ``Global-local bidirectional reasoning for unsupervised representation learning of 3d point clouds,'' in \emph{Proceedings of the IEEE Conference on Computer Vision and Pattern Recognition (CVPR)}, 2020, pp. 5376--5385.

\bibitem{te2018rgcnn}
G.~Te, W.~Hu, A.~Zheng, and Z.~Guo, ``Rgcnn: Regularized graph cnn for point cloud segmentation,'' in \emph{Proceedings of the 26th ACM international conference on Multimedia}, 2018, pp. 746--754.

\bibitem{gao2020graphter}
X.~Gao, W.~Hu, and G.-J. Qi, ``Graphter: Unsupervised learning of graph transformation equivariant representations via auto-encoding node-wise transformations,'' in \emph{Proceedings of the IEEE/CVF Conference on Computer Vision and Pattern Recognition}, 2020, pp. 7163--7172.

\bibitem{su2015multi}
H.~Su, S.~Maji, E.~Kalogerakis, and E.~Learned-Miller, ``Multi-view convolutional neural networks for 3d shape recognition,'' in \emph{Proceedings of the IEEE International Conference on Computer Vision (ICCV)}, 2015, pp. 945--953.

\bibitem{yu2018multi}
T.~Yu, J.~Meng, and J.~Yuan, ``Multi-view harmonized bilinear network for 3d object recognition,'' in \emph{Proceedings of the IEEE Conference on Computer Vision and Pattern Recognition (CVPR)}, 2018, pp. 186--194.

\bibitem{graham20183d}
B.~Graham, M.~Engelcke, and L.~Van Der~Maaten, ``3d semantic segmentation with submanifold sparse convolutional networks,'' in \emph{Proceedings of the IEEE Conference on Computer Vision and Pattern Recognition (CVPR)}, 2018, pp. 9224--9232.

\bibitem{wang2018sgpn}
W.~Wang, R.~Yu, Q.~Huang, and U.~Neumann, ``Sgpn: Similarity group proposal network for 3d point cloud instance segmentation,'' in \emph{Proceedings of the IEEE Conference on Computer Vision and Pattern Recognition (CVPR)}, 2018, pp. 2569--2578.

\bibitem{xu2020grid}
Q.~Xu, X.~Sun, C.-Y. Wu, P.~Wang, and U.~Neumann, ``Grid-gcn for fast and scalable point cloud learning,'' in \emph{Proceedings of the IEEE Conference on Computer Vision and Pattern Recognition (CVPR)}, 2020, pp. 5661--5670.

\bibitem{yang2019learning}
B.~Yang, J.~Wang, R.~Clark, Q.~Hu, S.~Wang, A.~Markham, and N.~Trigoni, ``Learning object bounding boxes for 3d instance segmentation on point clouds,'' \emph{arXiv preprint arXiv:1906.01140}, 2019.

\bibitem{li2018pointcnn}
Y.~Li, R.~Bu, M.~Sun, W.~Wu, X.~Di, and B.~Chen, ``Pointcnn: Convolution on x-transformed points,'' \emph{Advances in Neural Information Processing Systems (NIPS)}, vol.~31, pp. 820--830, 2018.

\bibitem{zaheer2017deep}
M.~Zaheer, S.~Kottur, S.~Ravanbakhsh, B.~Poczos, R.~R. Salakhutdinov, and A.~J. Smola, ``Deep sets,'' \emph{Advances in Neural Information Processing Systems (NIPS)}, vol.~30, 2017.

\bibitem{duan2019structural}
Y.~Duan, Y.~Zheng, J.~Lu, J.~Zhou, and Q.~Tian, ``Structural relational reasoning of point clouds,'' in \emph{Proceedings of the IEEE Conference on Computer Vision and Pattern Recognition (CVPR)}, 2019, pp. 949--958.

\bibitem{liu2019densepoint}
Y.~Liu, B.~Fan, G.~Meng, J.~Lu, S.~Xiang, and C.~Pan, ``Densepoint: Learning densely contextual representation for efficient point cloud processing,'' in \emph{Proceedings of the IEEE International Conference on Computer Vision (ICCV)}, 2019, pp. 5239--5248.

\bibitem{yang2019modeling}
J.~Yang, Q.~Zhang, B.~Ni, L.~Li, J.~Liu, M.~Zhou, and Q.~Tian, ``Modeling point clouds with self-attention and gumbel subset sampling,'' in \emph{Proceedings of the IEEE Conference on Computer Vision and Pattern Recognition (CVPR)}, 2019, pp. 3323--3332.

\bibitem{atzmon2018point}
M.~Atzmon, H.~Maron, and Y.~Lipman, ``Point convolutional neural networks by extension operators,'' \emph{arXiv preprint arXiv:1803.10091}, 2018.

\bibitem{thomas2019kpconv}
H.~Thomas, C.~R. Qi, J.-E. Deschaud, B.~Marcotegui, F.~Goulette, and L.~J. Guibas, ``Kpconv: Flexible and deformable convolution for point clouds,'' in \emph{Proceedings of the IEEE International Conference on Computer Vision (ICCV)}, 2019, pp. 6411--6420.

\bibitem{simonovsky2017dynamic}
M.~Simonovsky and N.~Komodakis, ``Dynamic edge-conditioned filters in convolutional neural networks on graphs,'' in \emph{Proceedings of the IEEE Conference on Computer Vision and Pattern Recognition (CVPR)}, 2017, pp. 3693--3702.

\bibitem{shen2018mining}
Y.~Shen, C.~Feng, Y.~Yang, and D.~Tian, ``Mining point cloud local structures by kernel correlation and graph pooling,'' in \emph{Proceedings of the IEEE Conference on Computer Vision and Pattern Recognition (CVPR)}, 2018, pp. 4548--4557.

\bibitem{du2021self}
B.~Du, X.~Gao, W.~Hu, and X.~Li, ``Self-contrastive learning with hard negative sampling for self-supervised point cloud learning,'' in \emph{Proceedings of the 29th ACM International Conference on Multimedia}, 2021, pp. 3133--3142.

\bibitem{zhao2021point}
H.~Zhao, L.~Jiang, J.~Jia, P.~H. Torr, and V.~Koltun, ``Point transformer,'' in \emph{Proceedings of the IEEE/CVF International Conference on Computer Vision}, 2021, pp. 16\,259--16\,268.

\bibitem{ma2022rethinking}
X.~Ma, C.~Qin, H.~You, H.~Ran, and Y.~Fu, ``Rethinking network design and local geometry in point cloud: A simple residual mlp framework,'' in \emph{International Conference on Learning Representations (ICLR)}, 2022.

\bibitem{yin2021graph}
J.~Yin, J.~Shen, X.~Gao, D.~Crandall, and R.~Yang, ``Graph neural network and spatiotemporal transformer attention for 3d video object detection from point clouds,'' \emph{IEEE Transactions on Pattern Analysis and Machine Intelligence}, 2021.

\bibitem{moosavi2016deepfool}
S.-M. Moosavi-Dezfooli, A.~Fawzi, and P.~Frossard, ``Deepfool: a simple and accurate method to fool deep neural networks,'' in \emph{Proceedings of the IEEE Conference on Computer Vision and Pattern Recognition (CVPR)}, 2016, pp. 2574--2582.

\bibitem{moosavi2017universal}
S.-M. Moosavi-Dezfooli, A.~Fawzi, O.~Fawzi, and P.~Frossard, ``Universal adversarial perturbations,'' in \emph{Proceedings of the IEEE Conference on Computer Vision and Pattern Recognition (CVPR)}, 2017, pp. 1765--1773.

\bibitem{mustafa2020deeply}
A.~Mustafa, S.~H. Khan, M.~Hayat, R.~Goecke, J.~Shen, and L.~Shao, ``Deeply supervised discriminative learning for adversarial defense,'' \emph{IEEE transactions on pattern analysis and machine intelligence}, vol.~43, no.~9, pp. 3154--3166, 2020.

\bibitem{liu2023robust}
D.~Liu, W.~Hu, and X.~Li, ``Robust geometry-dependent attack for 3d point clouds,'' \emph{IEEE Transactions on Multimedia}, 2023.

\bibitem{tao20233dhacker}
Y.~Tao, D.~Liu, P.~Zhou, Y.~Xie, W.~Du, and W.~Hu, ``3dhacker: Spectrum-based decision boundary generation for hard-label 3d point cloud attack,'' in \emph{Proceedings of the IEEE/CVF International Conference on Computer Vision}, 2023, pp. 14\,340--14\,350.

\bibitem{liu2021imperceptible}
D.~Liu and W.~Hu, ``Imperceptible transfer attack and defense on 3d point cloud classification,'' \emph{arXiv preprint arXiv:2111.10990}, 2021.

\bibitem{zhao2020nudge}
Y.~Zhao, I.~Shumailov, R.~Mullins, and R.~Anderson, ``Nudge attacks on point-cloud dnns,'' \emph{arXiv preprint arXiv:2011.11637}, 2020.

\bibitem{sun2021local}
Y.~Sun, F.~Chen, Z.~Chen, and M.~Wang, ``Local aggressive adversarial attacks on 3d point cloud,'' in \emph{Asian Conference on Machine Learning}.\hskip 1em plus 0.5em minus 0.4em\relax PMLR, 2021, pp. 65--80.

\bibitem{liao2018defense}
F.~Liao, M.~Liang, Y.~Dong, T.~Pang, X.~Hu, and J.~Zhu, ``Defense against adversarial attacks using high-level representation guided denoiser,'' in \emph{Proceedings of the IEEE conference on computer vision and pattern recognition}, 2018, pp. 1778--1787.

\bibitem{jia2019comdefend}
X.~Jia, X.~Wei, X.~Cao, and H.~Foroosh, ``Comdefend: An efficient image compression model to defend adversarial examples,'' in \emph{Proceedings of the IEEE/CVF conference on computer vision and pattern recognition}, 2019, pp. 6084--6092.

\bibitem{zhou2019dup}
H.~Zhou, K.~Chen, W.~Zhang, H.~Fang, W.~Zhou, and N.~Yu, ``Dup-net: Denoiser and upsampler network for 3d adversarial point clouds defense,'' in \emph{Proceedings of the IEEE International Conference on Computer Vision (ICCV)}, 2019, pp. 1961--1970.

\bibitem{dong2020self}
X.~Dong, D.~Chen, H.~Zhou, G.~Hua, W.~Zhang, and N.~Yu, ``Self-robust 3d point recognition via gather-vector guidance,'' in \emph{Proceedings of the IEEE Conference on Computer Vision and Pattern Recognition (CVPR)}, 2020, pp. 11\,513--11\,521.

\bibitem{wu2020if}
Z.~Wu, Y.~Duan, H.~Wang, Q.~Fan, and L.~J. Guibas, ``If-defense: 3d adversarial point cloud defense via implicit function based restoration,'' \emph{arXiv preprint arXiv:2010.05272}, 2020.

\bibitem{liu2021pointguard}
H.~Liu, J.~Jia, and N.~Z. Gong, ``Pointguard: Provably robust 3d point cloud classification,'' in \emph{Proceedings of the IEEE Conference on Computer Vision and Pattern Recognition (CVPR)}, 2021, pp. 6186--6195.

\bibitem{rosman2013patch}
G.~Rosman, A.~Dubrovina, and R.~Kimmel, ``Patch-collaborative spectral point-cloud denoising,'' in \emph{Computer Graphics Forum}, vol.~32, no.~8.\hskip 1em plus 0.5em minus 0.4em\relax Wiley Online Library, 2013, pp. 1--12.

\bibitem{zhang2020hypergraph}
S.~Zhang, S.~Cui, and Z.~Ding, ``Hypergraph spectral analysis and processing in 3d point cloud,'' \emph{IEEE Transactions on Image Processing}, vol.~30, pp. 1193--1206, 2020.

\bibitem{chen2017fast}
S.~Chen, D.~Tian, C.~Feng, A.~Vetro, and J.~Kova{\v{c}}evi{\'c}, ``Fast resampling of three-dimensional point clouds via graphs,'' \emph{IEEE Transactions on Signal Processing}, vol.~66, no.~3, pp. 666--681, 2017.

\bibitem{ramasinghe2020spectral}
S.~Ramasinghe, S.~Khan, N.~Barnes, and S.~Gould, ``Spectral-gans for high-resolution 3d point-cloud generation,'' in \emph{2020 IEEE/RSJ International Conference on Intelligent Robots and Systems (IROS)}.\hskip 1em plus 0.5em minus 0.4em\relax IEEE, 2020, pp. 8169--8176.

\bibitem{chung1997spectral}
F.~R. Chung and F.~C. Graham, \emph{Spectral graph theory}.\hskip 1em plus 0.5em minus 0.4em\relax American Mathematical Soc., 1997, no.~92.

\bibitem{xu2019predictive}
Y.~Xu, W.~Hu, S.~Wang, X.~Zhang, S.~Wang, S.~Ma, Z.~Guo, and W.~Gao, ``Predictive generalized graph {F}ourier transform for attribute compression of dynamic point clouds,'' \emph{arXiv preprint arXiv:1908.01970}, 2019.

\bibitem{wu20153d}
Z.~Wu, S.~Song, A.~Khosla, F.~Yu, L.~Zhang, X.~Tang, and J.~Xiao, ``3d shapenets: A deep representation for volumetric shapes,'' in \emph{Proceedings of the IEEE Conference on Computer Vision and Pattern Recognition (CVPR)}, 2015, pp. 1912--1920.

\bibitem{fan2017point}
H.~Fan, H.~Su, and L.~J. Guibas, ``A point set generation network for 3d object reconstruction from a single image,'' in \emph{Proceedings of the IEEE Conference on Computer Vision and Pattern Recognition (CVPR)}, 2017, pp. 605--613.

\bibitem{huttenlocher1993comparing}
D.~P. Huttenlocher, G.~A. Klanderman, and W.~J. Rucklidge, ``Comparing images using the hausdorff distance,'' \emph{IEEE Transactions on Pattern Analysis and Machine Intelligence}, vol.~15, no.~9, pp. 850--863, 1993.

\bibitem{wang2022art}
R.~Wang, Y.~Yang, and D.~Tao, ``Art-point: Improving rotation robustness of point cloud classifiers via adversarial rotation,'' in \emph{Proceedings of the IEEE/CVF Conference on Computer Vision and Pattern Recognition}, 2022, pp. 14\,371--14\,380.

\bibitem{kingma2014adam}
D.~P. Kingma and J.~Ba, ``Adam: A method for stochastic optimization,'' \emph{arXiv preprint arXiv:1412.6980}, 2014.

\bibitem{cortes2012l2}
C.~Cortes, M.~Mohri, and A.~Rostamizadeh, ``L2 regularization for learning kernels,'' \emph{arXiv preprint arXiv:1205.2653}, 2012.

\bibitem{liu2022boosting}
B.~Liu, J.~Zhang, and J.~Zhu, ``Boosting 3d adversarial attacks with attacking on frequency,'' \emph{IEEE Access}, vol.~10, pp. 50\,974--50\,984, 2022.

\bibitem{huang2022shape}
Q.~Huang, X.~Dong, D.~Chen, H.~Zhou, W.~Zhang, and N.~Yu, ``Shape-invariant 3d adversarial point clouds,'' in \emph{Proceedings of the IEEE/CVF Conference on Computer Vision and Pattern Recognition}, 2022, pp. 15\,335--15\,344.

\bibitem{li2022robust}
K.~Li, Z.~Zhang, C.~Zhong, and G.~Wang, ``Robust structured declarative classifiers for 3d point clouds: Defending adversarial attacks with implicit gradients,'' in \emph{Proceedings of the IEEE/CVF Conference on Computer Vision and Pattern Recognition}, 2022, pp. 15\,294--15\,304.

\bibitem{sun2022pointdp}
J.~Sun, W.~Nie, Z.~Yu, Z.~M. Mao, and C.~Xiao, ``Pointdp: Diffusion-driven purification against adversarial attacks on 3d point cloud recognition,'' \emph{arXiv preprint arXiv:2208.09801}, 2022.

\end{thebibliography}

\vspace{-40pt}
\begin{IEEEbiography}
 [{\includegraphics[width=1in,height=1.25in,clip,keepaspectratio] {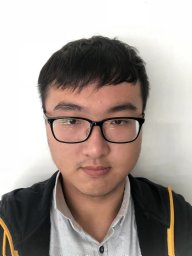}}]
 {Daizong Liu}
 received the B.S. degree in Information Engineering from Wuhan University of Technology in 2018 and the M.S. degree in Electronic Information and Communication of Huazhong University of Science and Technology in 2021. He is currently working toward the Ph.D. degree at Wangxuan Institute of Computer Technology of Peking University.
 His research interests include 3D adversarial attacks, multi-modal learning, etc.
\end{IEEEbiography}

\vspace{-40pt}
\begin{IEEEbiography}
 [{\includegraphics[width=1in,height=1.25in,clip,keepaspectratio] {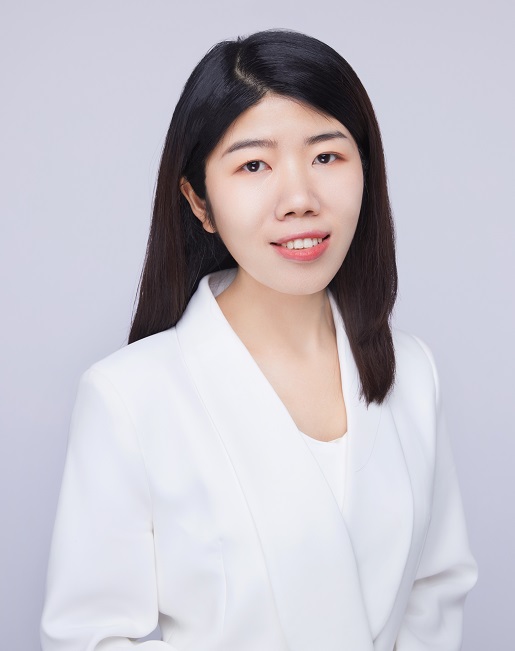}}]
 {Wei Hu}
 (M'17-SM'21) received the B.S. degree in Electrical Engineering from the University of Science and Technology of China in 2010, and the Ph.D. degree in Electronic and Computer Engineering from the Hong Kong University of Science and Technology in 2015.
 She was a Researcher with Technicolor, Rennes, France, from 2015 to 2017. 
 She is currently an Associate Professor with Wangxuan Institute of Computer Technology, Peking University. Her research interests are graph signal processing, graph-based machine learning and 3D visual computing. 
 She was awarded the 2021 IEEE Multimedia Rising Star Award---Honorable Mention, and received several paper awards including Best Paper Candidate at CVPR 2021 and Best Student Paper Runner Up Award at ICME 2020.    
 She serves as an Associate Editor for Signal Processing Magazine and IEEE Transactions on Signal and Information Processing over Networks. 
\end{IEEEbiography}

\vspace{-40pt}
\begin{IEEEbiography}[{\includegraphics[width=1in,height=1.25in,clip,keepaspectratio]{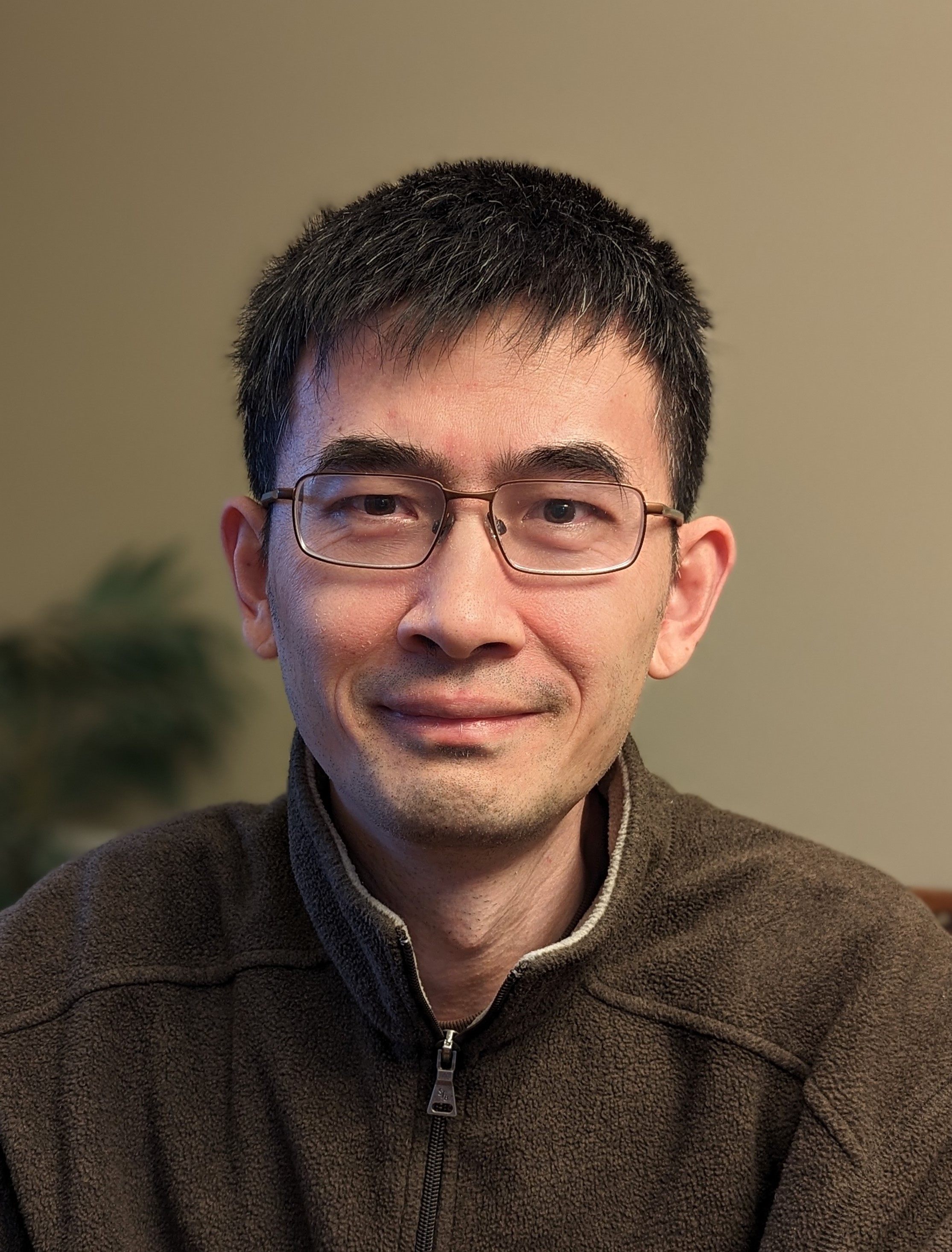}}]{Xin Li}
		received the B.S. degree with highest honors in electronic engineering and information science from University of Science and Technology of China, Hefei, in 1996, and the Ph.D. degree in electrical engineering from Princeton University, Princeton, NJ, in 2000. He was a Member of
Technical Staff with Sharp Laboratories of America, Camas, WA from Aug. 2000 to Dec. 2002. 
He was a faculty member in Lane Department of Computer Science and Electrical Engineering, West Virginia University from Jan. 2003 to Aug. 2023. 
Currently, he is with the Department of Computer Science, University at Albany, Albany, NY 12222 USA. His research interests include image and video processing, compute vision and computational neuroscience. 
Dr. Li was elected a Fellow of IEEE in 2017 for his contributions to image interpolation, restoration and compression.

\end{IEEEbiography}

\end{document}